\documentclass[journal, final]{IEEEtran}
\usepackage{times}

\usepackage[numbers]{natbib}
\usepackage{multicol}
\usepackage[bookmarks=true]{hyperref}

\usepackage{graphicx}
\usepackage{amsmath}
\usepackage{amssymb}
\usepackage{booktabs}
\usepackage{subfig}

\makeatletter  
\newif\if@restonecol  
\makeatother

\usepackage[linesnumbered,ruled,vlined]{algorithm2e}%[ruled,vlined]{  
\usepackage{algpseudocode}  
\usepackage{amsmath}  
\newcommand{\noteCM}{\textcolor{blue}}

\usepackage[capitalize]{cleveref}
\crefname{section}{Sec.}{Secs.}
\Crefname{section}{Section}{Sections}
\Crefname{table}{Table}{Tables}
\crefname{table}{Tab.}{Tabs.}

\usepackage[dvipsnames]{xcolor}
\definecolor{myblue}{RGB}{164, 180, 255}
\definecolor{mygreen}{RGB}{100, 200, 10}
\definecolor{mypink}{RGB}{244, 204, 204}

\usepackage{adjustbox}
\usepackage{multirow}
\usepackage{float}

\newcommand{\ours}{TNS}

\begin{document}

% paper title
\title{TNS: Terrain Traversability Mapping and Navigation System for Autonomous Excavators}

% You will get a Paper-ID when submitting a pdf file to the conference system
% \author{Author Names Omitted for Anonymous Review. Paper-ID 64 \thanks{test}}

\author{
Tianrui Guan$^{1*}$\thanks{* Work done during an internship at Baidu RAL.} %
% {\tt\small rayguan@terpmail.umd.edu}
~~~Zhenpeng He$^{1}$
% {\tt\small hezhp@shanghaitech.edu.cn}
~~~Ruitao Song$^{1}$
% {\tt\small xxx@xxxx}
~~~Dinesh Manocha$^{2}$
% {\tt\small dmanocha@umd.edu}
~~~Liangjun Zhang$^{1}$\\
% {\tt\small liangjun.zhang@gmail.com}\\
% \and
$^1$ Robotics and Auto-Driving Laboratory, Baidu Research\\
$^2$ University of Maryland, College Park \\
% \small{(Full report, video and dataset at \url{https://gamma.umd.edu/ttm})}
}

\maketitle

%%%%%%%%% ABSTRACT
\begin{abstract}

We present a terrain traversability mapping and navigation system (\ours) for autonomous excavator applications in an unstructured environment. We use an efficient approach to extract terrain features from RGB images and 3D point clouds and incorporate them into a global map for planning and navigation. Our system can adapt to changing environments and update the terrain information in real-time. Moreover, we present a novel dataset, the Complex Worksite Terrain (CWT) dataset, which consists of RGB images from construction sites with seven categories based on navigability. Our novel algorithms improve the mapping accuracy over previous SOTA methods by $4.17-30.48\%$ and reduce MSE on the traversability map by $13.8-71.4\%$. We have combined our mapping approach with planning and control modules in an autonomous excavator navigation system and observe $49.3\%$ improvement in the overall success rate. Based on \ours{}, we demonstrate the first autonomous excavator that can navigate through unstructured environments consisting of deep pits, steep hills, rock piles, and other complex terrain features. Dataset, videos, and a full technical report are available at \href{https://gamma.umd.edu/tns/}{gamma.umd.edu/tns/}.

\end{abstract}

%%%%%%%%% BODY TEXT

\IEEEpeerreviewmaketitle

% \vspace{-10pt}
\section{Introduction}

Excavators are one of the most common types of heavy-duty machinery used for earth-moving activities, including mining, construction, environmental restoration, etc. 
% According to~\cite{Zhangeabc3164}, the size of the global market share for excavators had reached \$44.12 billion in 2018 and is expected to grow to \$63.14 billion by 2026. 
As the demand for excavators increases, many autonomous excavator systems~\cite{Zhangeabc3164, aes2003, aes2011} have been proposed for material loading tasks, which involve perception and motion planning techniques. 
%Autonomous excavator system (AES)~\cite{Zhangeabc3164} is the first deployed system to perform loading tasks in real-world scenarios.

%In this paper, we address some of the challenges related to perception and its impact on navigation for autonomous excavator.

Some of the major issues in terms of using autonomous excavators are the development of robust perception and navigation sub-systems.
In general, perception in unstructured environments such as excavation  has many challenges. There have been many works related to unstructured environments, including perception and terrain classification~\cite{guan2021ganav, viswanath2021offseg, singh2021offroadtranseg} and navigation~\cite{terrain_class_nav_2005, kahn2020badgr, small_nav1, kumar2021rma}. Applications in unstructured, hazardous environments have even more difficulties in terms of  robustness and limitations on the computational budget. For example, many accurate learning methods have been proposed to improve the perception capabilities, but we cannot assume access to large GPUs or clusters for excavators operating in hazardous environments. Instead, we need to develop robust methods with lower computational requirements.
%Even though there has many datasets~\cite{RUGD2019IROS, rellis} trying to tackle similar problem as a computer vision tasks, there is still a gap between good perception results and good application outcomes.

\begin{figure}[t]
    \centering
    \includegraphics[width = \columnwidth]{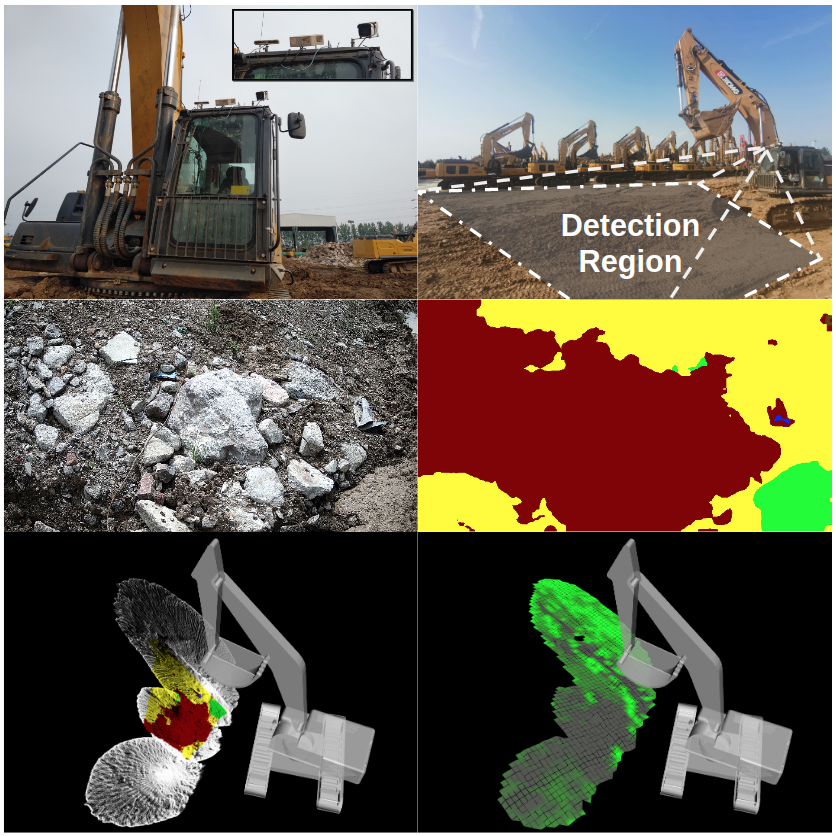}
    \vspace{-10pt}
    \caption{Overview of our system \ours{}: 
    % TTM is the first approach for excavator navigation in complex, unstructured environments. 
    \textbf{Top left:} Sensors on the excavator, including RGB cameras and Livox LiDAR. 
    % GPS RTK is not visible in this figure. 
    \textbf{Top right:} Detection region from a third-person perceptive. \textbf{Middle left:} Frontal view captured by the camera. \textbf{Middle right:} Semantic segmentation output, where green, yellow, and maroon correspond to flat region, bumpy region, and rock, respectively. \textbf{Bottom left:} Colored point cloud with semantic labels. \textbf{Bottom right:} Terrain traversability output, where the traversability value decreases from green to grey. This output is used for automatic navigation in complex, outdoor environments.}
    % \vspace{-15pt}
    \vspace{-8mm}
    \label{fig:cover}
\end{figure}

Traversability is a term that encompasses both perception and navigation. It has been well-studied for decade, and there have been many works~\cite{geo1, pc_desert_cmp1, semantic_for_planning_cmp2, 3d_cam_cmp3, semantic_only_cmp4} on traversability estimation for planning and navigation. Terrain traversability is a binary value, or a probability score, measuring the difficulty of navigating a region through perception sensors like camera, LiDAR and IMU. Terrain traversability estimation is a critical step between perception and navigation. In many autonomous driving (AD) cases~\cite{ small_nav1, lidar_obstacle}, a method capable of detecting obstacles and distinguishing road and non-road regions is sufficient for navigation.  On the other hand, in an unstructured, hazardous environment where off-road navigation is unavoidable, there are many factors that must be considered, including efficiency, adaptability, and safety. In such cases, not only a more detailed classification according to terrain features is needed, but also a continuous value for traversability is preferred to describe the complexity of the terrain and provide the best option for the navigation module. Therefore, we need good techniques to detect traversable regions for reliable navigation in an unstructured scene.

\noindent{\bf Main Results:}
%In this paper we aim to discuss how to tackle navigation problems in unstructured environment through a good map representation from a perception perspective, and highlight effectiveness of our method through different measurements, from accuracy of the map, to planning performance and navigation demonstration on an excavator.
We present a terrain traversability mapping and navigation system (\ours) for  traversability classification and autonomous navigation. We describe an efficient semantic-geometric fusion method to extract traversability maps. Our method leverages the physical and computational constraints of the robot, including maximum climbing degree, width of the body, run-time computational budget, etc. The novel aspects of our approach include:

\begin{enumerate}

  %  \item We discuss several perception challenges in unstructured environments and present some good measurements and criterion in terms of perception and mapping for excavator application. 
    
    \item We present a real-time terrain traversability estimation and navigation system (\ours{}) from 3D LiDAR and RGB camera inputs for mapping, planning, and navigation. We describe a novel learning-based geometric fusion solution that considers machine specifications and hardware limitations for terrain traversability prediction in unstructured environments.
    % based on learning-and-geometric method as well as the machine specifications. 
    % We use learned semantic terrain information as visual guidance for dangerous surface detection and use geometric terrain features as supplemental information in uncertain and unknown regions.
    % \item A new LiDAR-camera fusion strategy for terrain traversability prediction in unstructured environment, based on learning-and-geometric method.
 We show that our method is the state-of-the-art (SOTA) traversability mapping method on complex terrains. Our method outperforms previous SOTA methods by 4.17-30.48\% in terms of mAcc and reduces the MSE by 13.8-71.4\%. 
    
    \item We have integrated \ours{} with planning and control algorithms and evaluated the performance extensively in real-world settings on an autonomous excavator in various challenging construction scenes, as shown in Figure~\ref{fig:cover}. We also elaborate on many non-trivial issues that came up during the implementation and evaluation and how we address them. We show that our \ours{} can safely navigate an excavator in unstructured environments and observe a 49\% improvement in terms of planning success rate. We highlight the benefits of \ours{} as the first autonomous excavator that can navigate through complex, unstructured environments.
    %The online implementation code is proprietary and will not be released.

    % This approach is integrated with planning and control modules and used to safely guide an excavator in unstructured environments.
    
    \item We present the Complex Worksite Terrain (CWT) dataset, which consists of 30 minutes of video and 669 RGB images in unstructured environments with seven different classes based on terrain types, traversable regions, and obstacles. We will release the CWT dataset in the public domain.
    % AET dataset will be released for future works on off-road perception.

% In addition, we present Complex Worksite Terrain (CWT) dataset, which includes a total of 30 minutes video and 669 RGB images from the working environments of excavators, annotated with seven categories according to surface navigability. CWT dataset provides us good traversability estimation on similar construction sites where excavators usually operate. 

\end{enumerate}
% We will release the implementation, terrain map outputs, and evaluation code that can reproduce the performance. The code and dataset link is shared in the supplemental material.

\section{Related Work}

\subsection{Field Robots and Systems}

Field robots usually refer to machines that operate in off-road, hazardous environments. 
These include heavy-duty service robots for industrial usage in mining~\cite{mining2019}, excavation~\cite{Zhangeabc3164}, agriculture~\cite{agricultural2018}, construction~\cite{construction_2020}, etc. 
% Computer vision can solve robotic problems in a variety of industrial usages, including mining~\cite{mining2019}, excavation~\cite{Zhangeabc3164}, agriculture~\cite{agricultural2018}, construction~\cite{construction_2020}, etc. 
To satisfy industrial needs and save labor costs, many automated systems~\cite{Zhangeabc3164, aes2003, aes2011} have been developed for service robots in the field. These systems include modules for perception, planning and control. However, it remains a challenge to fully automate many tasks in unknown, unstructured environments.

\subsection{Terrain Traversability Recognition}

% The concept of traversability has been in discussion for decades, which are also referred as "drivability", "navigability", etc~\cite{survey1}. 
The concept of traversability, also referred to as ``drivability," ``navigability," etc.~\cite{survey1}, has been studied for decades.
% Under the same theme, they have different view points of such problem and investigate such topic with different evaluation methods and goals. 
There are many viewpoints on the problems and challenges associated with traversability, and investigations into such topics have had different evaluation methods and goals.
Many works focus on getting correct predictions of the terrain~\cite{sift, material, rgb1, rgb2, rgb3, binary1, simulation_binary2, guan2021ganav, gonet, 2d_lane_det} by some notion of ground truth based on human-labeled annotation, similar to the metrics of 2D and 3D semantic segmentation. Most of the methods mentioned above are based on visual features of the terrain, which sometimes lack the properties that enable real-world navigation due to recognition failure.

On the other hand, some works focus on obtaining traversability maps that result in the best navigation outcomes. There are plenty of works~\cite{semantic_mapping, semantic_only_cmp4, trav_cost, pc_desert_cmp1, cp_structured, semantic_for_planning_cmp2} on classifying different terrains based on either material categories or navigability properties and demonstrate their mapping results through navigation outcomes. 

However, those methods deal with structured roads or roads with clear path boundaries in unstructured environments. In more complex environments, point clouds obtained from LiDAR are used to extract geometric attributes of the surface, including slope, height variation, roughness, obstacles, etc., as proposed in \cite{geo1, geo_simple, geo2, geo_trav, lidar_nav, lidar_obstacle}. \cite{2D3D} uses both point cloud and RGB images to classify terrains with safe, risky, and obstacle labels in the 2D image plane for better performance. \cite{hybrid_camera_lidar} presents a pipeline from perception to motion control and uses five different data sources for navigation, including range and intensity values from a 2D LiDAR and edge information from an RGB-D camera. \cite{safe1, safe2} analyze the terrain and create roadmaps for road safety. The works most similar to our proposed method are \cite{pc_desert_cmp1, semantic_for_planning_cmp2, 3d_cam_cmp3, semantic_only_cmp4}, which focus on finding a better terrain representation for navigation in unstructured terrains.

\subsection{Datasets for Unstructured Environments}

Most recent developments in perception tasks like object detection and semantic segmentation focus on urban driving scene datasets like KITTI~\cite{kitti}, Waymo~\cite{waymo}, etc., which achieve high accuracy in terms of average precision. On the other hand, unstructured scenes like the natural environment, construction sites, and complicated traffic scenarios are less explored, for two  primary reasons. First, there are fewer datasets with unstructured environments; second, perception and autonomous navigation in unstructured off-road environments are challenging due to unpredictability and diverse terrain types.

Recent efforts in off-road perception and navigation include RUGD~\cite{RUGD2019IROS} and RELLIS-3D~\cite{rellis}, which are semantic segmentation datasets collected from a robot navigating in off-road and natural environments. These datasets contain scenes like trails, forests, creeks, etc. 
% The Indian Driving Dataset~\cite{idd} is a traffic dataset collected in India, where most of the roads are not well-delineated and traffic conditions are unpredictable. In addition,
\cite{construction_dataset} is a construction dataset containing annotations of heavy-duty vehicles for detection, tracking, and activity classifications.

\section{Perception for Autonomous Excavators}
\label{def}

\begin{figure*}[t]
    \vspace{3pt}
    \centering
    \includegraphics[width=\textwidth]{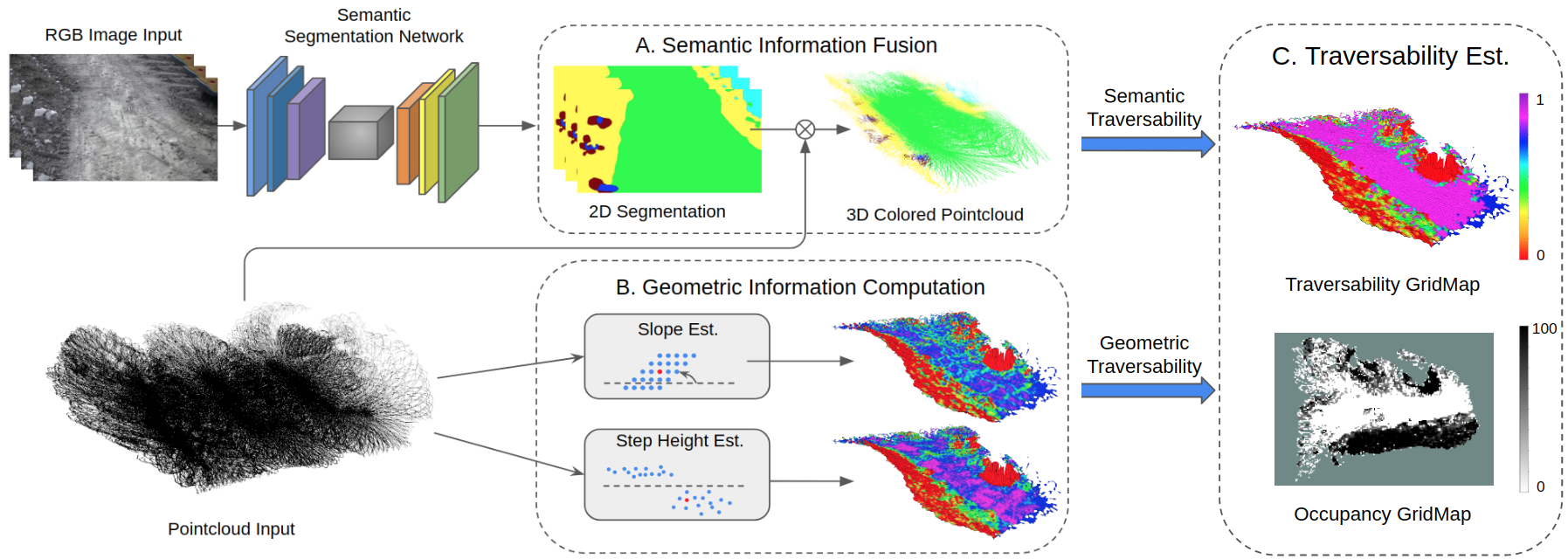}
    \caption{\textbf{Overview of the perception module in \ours{}:} Our system takes RGB images and point clouds as inputs to infer traversability. We extract semantic information using segmentation and associate terrain labels with point clouds, as shown in A (\textbf{top}). We extract geometric information using slope and step height estimation, as shown in B (\textbf{bottom}). We produce a traversability grid map based on semantic and geometric information and convert it to a 2D occupancy map for path planning and navigation, as shown in C (\textbf{right}).}
    \label{fig:overview}
    \vspace{-18pt}
\end{figure*}

The road conditions in structured environments such as highways are usually navigation-friendly, so the core problem during navigation in structured environments is avoiding obstacles rather than determining which part of the surface is easier and safer to navigate. 
% The boundaries of different objects in structured, urban environments can be clearly captured by perception methods, which reduces the likelihood of mis-classifications. 
In contrast, excavators are usually operated in unstructured and dangerous environments consisting of rock piles, cliffs, deep pits, steep hills, etc. 
% An unstructured environment usually refers to a terrain that lacks structure and has unpredictable and potentially hazardous conditions. 
Such an environment lacks any lane markings, and the arrangement of obstacles tends to be non-uniform. 
In addition, due to tasks like digging and dumping, the working conditions for excavators are constantly changing. Landfalls and cave-ins occur, potentially causing the excavator to tip over and injure the operator. Therefore, it is crucial to identify different terrains and predict safe regions for navigation. Furthermore, we need solutions with low computational requirements.

In our context, traversability~\cite{survey1} refers to the capability of a ground vehicle to reside over a region of terrain under an admissible state wherein it can enter given its current state. In order to solve navigation challenges for excavators as well as other working vehicles in unstructured terrain, we formulate the problem of obtaining an accurate traversability map representation as follows:

\noindent\textbf{Problem Definition:} Given sensor inputs $S_1, S_2, .., S_h$ from $h$ different sources over a time span $T$, the goal is to obtain a 2D grid map $T\in[0, 1]^{H \times W}$ with resolution $r$, where $T$ corresponds to some region $R$ of shape $(Hr, Wr)$. The maximum value corresponds to a non-traversable region and the minimum value corresponds to the most traversable region.

\noindent\textbf{Metrics for Traversability Map:} We need to consider the following measurements in excavator applications:
% \vspace{-6pt}
\begin{itemize}
    \item \textbf{Accuracy:} Similar to~\cite{2D3D, pc_desert_cmp1, 3d_cam_cmp3}, we use an ROC curve to measure the accuracy of the traversability prediction. In addition, the map output should fit the terrain closely, so we also use MSE (mean squared error) as a fitness measurement.
    \item \textbf{Performance:} \cite{semantic_only_cmp4, semantic_for_planning_cmp2} use navigation outcome to measure their terrain traversability mapping algorithms, which include travel time, success rate, etc.
    \item \textbf{Energy constraints and run-time:} Due to the limitations of hardware and power supply on the excavator, energy efficiency and run-time computational budget should also be measured in a terrain traversability mapping method.
\end{itemize}
\section{\ours{}: System Architecture}

\begin{figure}[t]
    \centering
    \includegraphics[width=\columnwidth]{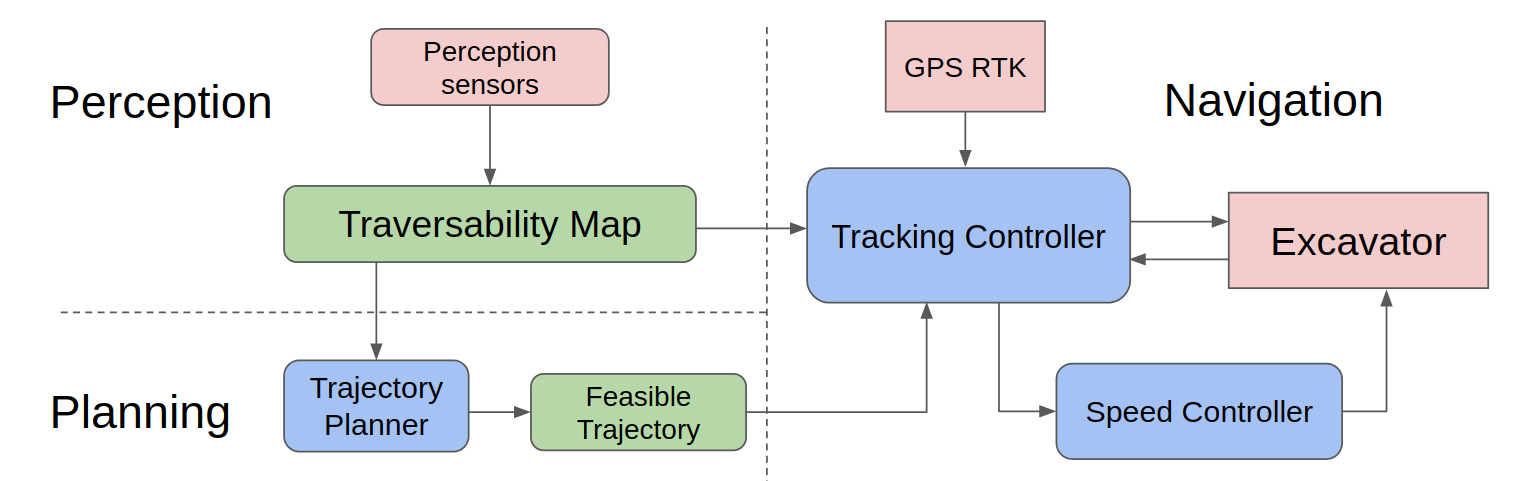}
    \caption{\textbf{Overall pipeline of \ours{} for autonomous excavator navigation:} We show different components of \ours{} as blue blocks and use green and red blocks to represent the intermediate output and hardware, respectively.}
    \label{fig:aes}
    \vspace{-15pt}
    % \vspace{-18pt}
\end{figure}

In this section, we describe our system for terrain traversability mapping and navigation (\ours{}) in excavator applications, as shown in Figure~\ref{fig:aes}. \ours{} takes a 3D point cloud stream from the LiDAR, an RGB camera stream from the RGB camera, and the corresponding poses of the excavator extracted from the GPS-RTK module. The goal of our proposed system is to identify safe, navigable regions for excavators and autonomously navigate the excavator based on the traversability map and the planned trajectory. The output of \ours{} includes a global map consisting of terrain information, including semantic information, geometric information, and a final traversability score, as well as the planned trajectory.

\subsection{Traversability Mapping}
\label{trav_mapping_A}

The terrain is represented as an elevation grid map and is updated in real-time based on incoming point clouds and RGB images.
Internally, each grid cell in the map stores the average height value of the latest $p$ points within this cell, as well as overall information about those points like update time, slope, step height, and their semantic information. A traversability score is calculated for each grid cell. In Figure~\ref{fig:overview}, we present an overview of our perception approach. Our implementation is based on the open-source grid map library~\cite{fankhauser2016universal}.

\noindent\textbf{Segmentation and Mapping to Point Cloud: }
We use 2D semantic segmentation on unstructured terrains. Given an input RGB image $I \in \mathbb{R}^{3\times H \times W}$,
% and the corresponding terrain labels $Y\in \mathbb{Z}^{H\times W}$,
the goal is to generate a mask $P\in\{0, 1, ..., N - 1\}^{H \times W}$, where $N$ is the number of classes. We use Fast-SCNN~\cite{fastscnn} after leveraging accuracy and efficiency, as shown in Table~\ref{tab:flops}.

% After we get the segmentation prediction $P$, we use a timestamp to locate the corresponding point cloud $C$. Given a point $\vec{p} = [x_p, y_p, z_p]\in C$ in the world coordinate, camera intrinsic matrix $K \in\mathbb{R}^{3\times 3}$, and extrinsic matrix $E\in\mathbb{R}^{4\times 4}$, we can calculate its corresponding image coordinates $\vec{p_i} = [u, v]$ based on the following formula:
% $$
% \begin{pmatrix}
% x_c \\
% y_c \\
% z_c \\
% 1
% \end{pmatrix}
%  = 
%  E \cdot 
% \begin{pmatrix}
% x_p \\
% y_p \\
% z_p \\
% 1
% \end{pmatrix} \\
% % \begin{pmatrix}
% % u \\
% % v \\
% % 1
% % \end{pmatrix}
% %  = 
% %  K \cdot
% % \begin{pmatrix}
% % x_c \\
% % y_c \\
% % z_c
% % \end{pmatrix}
% %  / z_c
% $$

% $$
% \hspace{15pt}
% \begin{pmatrix}
% u \\
% v \\
% 1
% \end{pmatrix}
%  = 
%  K \cdot
% \begin{pmatrix}
% x_c \\
% y_c \\
% z_c
% \end{pmatrix}
%  / z_c
% $$

After we get the segmentation prediction $P$, we use a timestamp to locate the corresponding point cloud $C$ and use camera calibration matrices to find the correspondence of each point to the segmentation results and save the terrain label in the grid map cell.

\noindent\textbf{Geometric Information Computation: }
In this section, we present details of slope and step height estimation and highlight how machine specifications are considered to calculate the geometric traversability score.

\subsubsection{Slope Estimation} Each grid cell $g$ is abstracted to a single point $p = \{x, y, z\}$, where $x$, $y$ is the center of the cell in the global coordinate frame and $z$ is the height value of the grid.
The slope $s$ in arbitrary grid cell $g$ is computed by the angle between the surface normal and the z-axis\footnote{Up direction in the real world} of the global coordinate frame:
$$
s = arcos({n}^z), {n}^z \in [0, 1]
$$
where ${n}^z$ is the component of normal $\vec{n}$ on the z-axis.

Similar to~\cite{geo1, mobile}, we use Principal Component Analysis (PCA) to calculate the normal direction of a grid cell. The covariance matrix $C_{cov}$ of the nearest neighbors of the query grid cell is calculated as follows:
\vspace{-3pt}
$$
C_{cov} = \frac{1}{k} \sum_{i=1}^{k} (p_i - \bar{p}) \cdot (p_i - \bar{p})^T , \ C_{cov} \cdot \vec{v_j} = \lambda_j \cdot \vec{v_j} , 
$$
$$
\ j \in \{0, 1, 2\}, \lambda_i < \lambda_j\ \text{if}\ i\ <\ j,
$$
where $k$ is the number of neighbors considered in the neighborhood of $g$, $p_i = \{x, y, z\}$ is the position of the neighbor grid in the global coordinate frame, $\bar{p}$ is the 3D centroid of the neighbors, $\lambda_j$ is the $j$-th eigenvalue of the covariance matrix, and $\vec{v_j}$ is the $j$-th eigenvector. The surface normal $\vec{n}$ of grid $g$ is the eigenvector $\vec{v_0}$ with the smallest absolute value of eigenvalue $\lambda_0$. 

% And we use the viewpoint $V = \{0, 0, 1\}$ to orient normal $\vec{n}$ consistently towards the viewpoint, they need to satisfy the equation: 
% $$
% \vec{n} \cdot V > 0
% $$

The purpose of the slope estimation is to get the shape of the terrain and avoid navigating on a steep surface. For excavator applications, the width between the tracks or wheels is a good indicator of the navigation stability on rough terrain. Usually, when the area of a rough region is less than half the width between the excavator's tracks, the excavator can navigate through it without any trouble. Specifically in our excavator setup, the width of our excavator track is $0.6\ m$, so we chose the grid resolution $d_{res} = 0.2\ m$ and search the nearest eight neighbors, which covers the necessary area.

\subsubsection{Step Height Estimation}
The step height $h$ is computed as the largest height difference between the center point $p$ of the grid and its $k'$ nearest neighbors: 
% \he{still a gap between g and p, should we use g instead}
$$
h = \max{ (\text{abs}(p^z - p_i^z))}, i \in [1,k']
$$
Since slope is a description of variation in the terrain in a relatively small region, we choose to use a neighbor search parameter $k' = 7*7 > k$ that spans $1.4\ m$ to measure height change in a larger scope. For excavator applications, the step height calculation guarantees that the track does not traverse areas with extreme height differences.

\subsubsection{Geometric Traversability Estimation}
Based on information about slope and step height of the terrain, we can calculate a geometric traversability score $T_{geo}$. According to the physical constraints of the robot, we create some critical values, $s_{cri}$, $s_{safe}$, $h_{cri}$, $h_{safe}$, as the thresholds for safety and danger detection. The purpose of those threshold values is to avoid danger when the surface condition exceeds the limits of the robot and to avoid more calculations when the surface is very flat. 
The formula for geometric traversability $T_{geo}$ for each grid is:

% The calculation of geometric traversability is determined based on the characteristics of the robot itself. 
% The track brings excellent terrain adaptability to the excavating robot. Therefore, the roughness of the terrain will not have a significant impact on the driving of the robot. 
% But the slope and step height are critical factors in determining the robot's traversability because of the high center of gravity.

% % \he{Give the specific value of the 490 document}
% In Fig. \ref{placeholder}, we can see that the slope is suitable for detecting dangerous areas such as pits or bumps. But we also notice some small areas(few numbers of cells) with high slopes on the map, usually caused by stones or pits on the ground. These areas should not be considered geometrically untraversable because they are too small to influence the robot. At this time, we need to use step height, which can be regarded as the integral of the slope, to judge those areas. So when the slope is large, but the step height is small, it should be treat as traversable. 
\vspace{-10pt}
\begin{equation}
\resizebox{.99 \hsize }{!}{$
T_{geo} = 
\begin{cases}
0& s > s_{cri} \ \text{or} \ h > h_{cri}\\
1& s < s_{safe} \ \text{and} \ h < h_{safe}\\
\max( 1 - (\alpha_1 \frac{s}{s_{cri}} + \alpha_2 \frac{h}{h_{cri}}) , 0 ) & \text{otherwise}
\end{cases},
$} \nonumber
\end{equation}
where the weights $\alpha_1$ and $\alpha_2$ sum up to 1. 

The step height estimation is complementary to slope estimation; it provides a global perspective, whereas slope is local terrain information. Combining these two specifications can help us remove noise in the map, such as bumps caused by dust, and ensure the robustness of the $T_{geo}$.

% For roughness, $r_{cri}$ is set to a heuristic value of $0.1\ m$ according to our experience that our robot can walk through steps less than $10\ cm$ high.

% \he{The result is not scale to 0~1!!!}
% roughness serve a a complimentary 

% \he{why we not use step height as the direct traversability but slope? noise? step height is the possible max influence of slope but can have some noise such as a pit and bump together?}
% \he{Only height values which were detected within a certain range $\delta f$ of frame numbers are considered in the traversability estimation process}

% \subsection{Traversability Estimation Combining Geometric and Semantic Information}
\noindent\textbf{Traversability with Geometric and Semantic Fusion: }
In this section, we describe our algorithm for geometric-semantic fusion. From the semantic and geometric information, we use a continuous traversability score $T\in [0, 1]$ to measure how easily the surface can be navigated. This is especially relevant to off-road scenarios because we prefer flat regions over bumpy roads to save energy. Moreover, when an excavator is navigating on a construction site, being able to correctly identify different regions is critical to avoid hazardous situations like flipping over.

The overall traversability score $T$ is calculated based on semantic terrain classes $C_{sem}$ and geometric traversability $T_{geo}$ on each grid:

% \begin{equation}
% \resizebox{.91 \hsize }{!}{$
% T = 
% \begin{cases}
% % 0& T_{sem} = 0 \ \text{or} \ T_{geo} = 0\\
% \alpha_{sem}\times T_{sem} + (1 - \alpha_{sem}) \times T_{geo} & T_{sem}\  exists\\
% (1-  \alpha_{penalty})\times T_{geo} + \alpha_{penalty} & \text{otherwise}
% \end{cases},
% $}
% \end{equation}
\vspace{-10pt}
\begin{equation}
\resizebox{.89 \hsize }{!}{$
T = 
\begin{cases}
0 & C_{sem} = \text{\{rock, excavator, obstacle, water\}}\\
1 & C_{sem} = \text{\{flat\}}\ \text{and}\ T_{geo} > 0\\
T_{geo} & \text{otherwise}\\
% T_{geo} & C_{sem} = \text{\{bumpy, mixture, unassigned\}}\\
% T_{geo} & \text{otherwise}
\end{cases},
$} \nonumber
\end{equation}

% \noindent where $\alpha_{sem}$ is the weight for semantic calculation and $\alpha_{penalty}$ is the penalty score without semantic information. In our case, we choose $\alpha_{sem} = 0.7$ and $\alpha_{penalty} = \tian{xx}$.

% \he{Based on our formulation, 
% we first make sure that if the terrain is considered non-traversable in either geometric calculation or semantic inference, the overall traversability is 0. 
% We want to make sure the robot should absolutely avoid dangerous regions from either geometric or semantic perspective. 
% Then, we use semantic information ${flat}$ to correct the possible noise in $T_{geo}$. 
% In other situations (\{bumpy, water, mixed\}), we choose to trust the geometric information. }

% we use a weighted sum of those two terms for the traversability estimation. It is possible that there is only geometric information, since LiDAR has a wider view than the RGB camera; in this case, we choose to add a penalty of 0.3 to the traversability due to lack of information.

% Based on our formulation, we make sure that the robot avoids any obstacles like rocks or excavators and forbidden regions like water. 
% We also consider flat regions as traversable regions with a score of 1. In other cases like bumpy, mixed, or unassigned regions, we set the score according to the geometric information.
This method is simple yet more effective than other comparably complicated fusion methods~\cite{3d_cam_cmp3, semantic_for_planning_cmp2}, as demonstrated in Section~\ref{evaluation} and Section~\ref{aes_imp}.

\subsection{Traversability-based Planning}
\label{planning}
% \tian{Might need help from Ruitao and Zhenpeng for planning and control part!! }

% \noindent\textbf{Post-processing:} We have a post-processing step on the traversability map before planning. We remove some non-traversable regions that satisfy all of the following criteria:
% \vspace{-7pt}
% \begin{itemize}
% %   \setlength\itemsep{-0.3em}
%     \item The region has a traversability value less than some occupied threshold $t_{occ}$.
%     \item The spans of the region along the x-axis and the y-axis satisfy both: 1) less than half the distance between two tracks of the excavator $d_{track}$ and 2) an average height less than the height $h_{cab}$, which is the distance from the bottom of the cabin to the ground.
% \end{itemize}
% \vspace{-7pt}

% After post-processing, we transform the traversability gridmap to a 2D occupancy gridmap as an input for path planning. We use the Hybrid A* planner~\cite{hybridA} to generate a path and send the trajectory to the motion controller, which guides the excavator to follow this trajectory. 

%%%%%%%%%%%%%%%%%%%%%%%%%%%%%%%%%%% OPTION 1 start
% This post-processing step is mostly to accommodate existing planners, since they do not work well and fail to plan a feasible trajectory with too many scattered, noisy regions. 
%%%%%%%%%%%%%%%%%%%%%%%%%%%%%%%%%%% OPTION 1 end

%%%%%%%%%%%%%%%%%%%%%%%%%%%%%%%%%%% OPTION 2 start

% \noindent\textbf{Planning:} 

We modify Hybrid A*~\cite{hybridA} to calculate a trajectory based on the traversability map output after the post-processing step. Hybrid A* is a global path planner based on a 2D occupancy grid map as an input for trajectory planning. The planner will generate a trajectory and send it to the motion controller, which guides the excavator to follow this trajectory.

% Although the excavators' tracks can be controlled independently for sharp turns, the tracks will damage the road surface and can get stuck in dirt when the turning radius is small. Therefore, the Hybrid A* algorithm is chosen over the A* algorithm in our application for a smoother trajectory. 
The traditional Hybrid A* algorithm only considers the traveling distance and certain driving maneuvers (such as reversing, turning, etc.), not the ground condition and traversability. As a result, the autonomous excavators can  be easily navigated to areas with low traversability in real-world applications with the traditional Hybrid A* planner. To solve the problem, we extend the Hybrid A* algorithm by introducing \ours{} and calculating the traversability cost. Specifically, we calculate the cost to the start of a vertex, which is the distance from the start state to the vertex with extra reversing or turning cost, and is weighted by the traversability value obtained from \ours{}. In the improved Hybrid A* algorithm, the cost to start $g(x)$ is increased by $k_{\ours{}}\cdot \delta l + g_{extra}$ when performing vertex expansion from the parent to the child vertices, where $\delta l$ is the distance between the two nodes, $g_{extra}$ is the extra penalty for reversing and turning, and $k_{\ours{}}$ is the traversability weighting factor calculated by:
\begin{equation}
k_{\ours{}} = \frac{k_t T_t A_t + k_u T_u A_u}{A_t + A_u},
\nonumber
\end{equation}
where $A_t$ and $A_u$ are the areas covered by the two tracks and between two tracks, respectively, of the excavator from the parent to the child vertices; $T_t$ and $T_u$ are the mean traversability value of areas $A_t$ and $A_u$; and $k_t$ and $k_u$ are two calibrating parameters.

%%%%%%%%%%%%%%%%%%%%%%%%%%%%%%%%%%% OPTION 2 end

\subsection{Control and Navigation}

% \noindent We briefly explain other components in AES framework:
% \vspace{-7pt}

\noindent The trajectory tracking controller is composed of a lateral trajectory tracking controller and a longitudinal speed controller: 

% The desired speed is chosen as a fixed value.

\begin{itemize}
%   \setlength\itemsep{-0.3em}
    % \item \textbf{Trajectory Planner:} We use Hybrid A*~\cite{hybridA} to calculate a trajectory based on the traversability map output from \ours{}.
    \item \textbf{Tracking Controller:} This module can adjust the steering of the robot for path following. It outputs the desired steering rate based on the heading error and the cross-track error. The cross-track error is defined as the distance between the point on the path closest to the reference point of the excavator. The control commands for the left and right tracks of the excavator are calculated using a lookup table according to the speed proportional and integral (PI) error metric and the desired steering rate. The tracking controller is developed based on \cite{hoffmann2007autonomous}. 

    \item \textbf{Speed Controller:} This module can adjust the speed of the robot. The speed controller receives the actual speed from the sensor and calculates the PI error metric according to the desired speed. 
\end{itemize}
% \vspace{-7pt}
%Please refer to AES~\cite{Zhangeabc3164} for more details. 
% More results based on \ours{}-integrated planner and its benefits are given in Section~\ref{aes_imp}.

%  It considers the heading error and the cross-track error. The cross-track error is defined as the distance between the closest point on the path with reference point of the excavator. 

% traversability map ---> occ map ---> remove small block ---> planning ---> controller ----> move

\subsection{Benefits over Prior Methods}
% \tian{Need to add this part somewhere!!! see commented part below}
% Different from \cite{geo1}, we exclude the calculation of roughness in our formulation since excavators are designed to handle relatively rough terrains. In addition, roughness can be partially described either through slope and step height, or be captured by visual features from the RGB images. 
Previous perception methods for traversability calculation only use geometric approaches~\cite{geo1, geo2, mobile, mobile2} in simple scenarios for mobile robot applications, or they can only navigate in an off-road environment with a clear visual path~\cite{pc_desert_cmp1, semantic_for_planning_cmp2, 3d_cam_cmp3, semantic_only_cmp4}. Our system is the first one to focus on excavator navigation applications in very challenging environments consisting of pits, hills, rock piles, etc. without a clear pathways.
In addition, our experiments and data are based on real-world scenarios in a construction site. 
Our method also adapts to the physical constraints of excavators to determine threshold, resolution of the grid, and $k$ neighbors.

We test our system \ours{} on an excavator based on the Autonomous Excavator System~\cite{Zhangeabc3164}.
% The AES system has a pipeline of perception, planning, and navigation modules. 
% We integrate and evaluate our system through planning results and real-world navigation performance by using our mapping output, as shown in Figure~\ref{fig:aes}. 
Note that previous AES systems mainly focus on digging tasks, while our system focuses on providing accurate mapping estimation and navigation in unstructured environments.

\vspace{-5pt}
\section{Complex Worksite Terrain (CWT) Dataset}
\label{dataset}

\begin{figure}[t]
    \centering
    \includegraphics[width=\columnwidth]{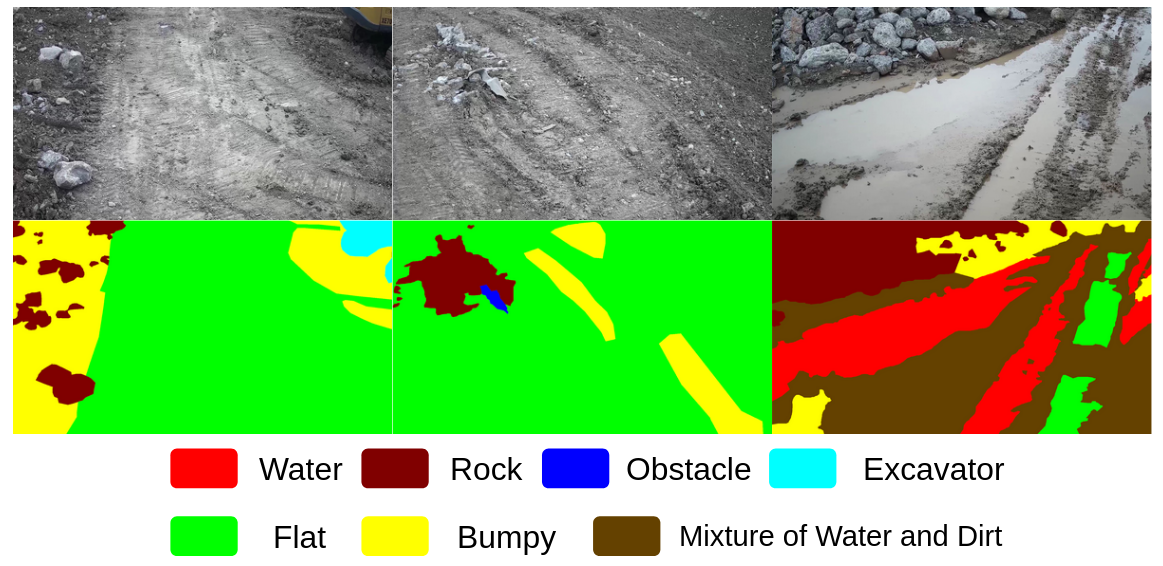}
    \vspace{-10pt}
    \caption{\textbf{Complex Worksite Terrain (CWT) dataset:} We show a few samples from our CWT dataset (\textbf{top}) and corresponding annotations (\textbf{bottom}). All images are collected in unstructured environments with various terrain types. }
    \label{fig:dataset_pic}
    % \vspace{-16pt}
    \vspace{-18pt}
\end{figure}

In this section, we present the Complex Worksite Terrain (CWT) dataset, which is collected at a construction site while an excavator is navigating through the work area. The hardware has the same setup as described in Section~\ref{hardware}.
We collect three videos (30 minutes in total) under different circumstances and annotate 669 images of size $1920\times1080$ according to terrain semantics. We only highlight the ontology and differences between CWT and other off-road datasets~\cite{rellis, RUGD2019IROS}, and provide details of the collection, class distribution, and analysis in the supplemental material.
% The training and testing sets are split according to a 9:1 ratio. 

% The CWT dataset is annotated with seven labels based on terrain features and navigability: flat region, bumpy region, mixture of water and dirt, water, rock, obstacle, and other working vehicles like excavators. The annotation is decided based on the opinion of a team of excavator operators. Based on their experience, flat surfaces, bumpy surfaces, and mixtures of water and dirt are usually navigable. In most cases, when flat surfaces are detected, they are preferable to other surfaces. Water is a forbidden region, since it is difficult to gauge how deep the water is, and the soil near the water could be soft and easily deformed. 

% Mark this part TBD
\begin{table}[b]
\centering
\vspace{-8 pt}
\resizebox{\columnwidth}{!}{%
\begin{tabular}{|c|c|c|c|}
\hline
Types                                                                 & Descriptions                                                                                                                               & Navigability & Distribution \\ \hline
\begin{tabular}[c]{@{}c@{}}Flat\\ Region\end{tabular}                 & \begin{tabular}[c]{@{}c@{}}Flat surfaces that most vehicles \\ like cars can traverse.\end{tabular}                                       & Easy         & 41.76\%      \\ \hline
\begin{tabular}[c]{@{}c@{}}Bumpy\\ Region\end{tabular}                & \begin{tabular}[c]{@{}c@{}}Bumpy surfaces that most vehicles \\ can not traverse except working \\ vehicles like excavators.\end{tabular} & Medium       & 42.59\%      \\ \hline
Rock Pile                  & \begin{tabular}[c]{@{}c@{}}Very common on work-site;\\ Need to be avoided most of the time.\end{tabular}                                   & Forbidden    & 6.51\%       \\ \hline
Water                                                                 & \begin{tabular}[c]{@{}c@{}}Water might be trapped in deep trench \\ after raining; Need to be avoided.\end{tabular}                       & Forbidden    & 3.66\%       \\ \hline
\begin{tabular}[c]{@{}c@{}}Mixtures of \\ Water and Dirt\end{tabular} & \begin{tabular}[c]{@{}c@{}}Shallow water with mostly\\ visible soil or dirt; Can be traversed.\end{tabular}                               & Medium       & 4.90\%       \\ \hline
\begin{tabular}[c]{@{}c@{}}Excavator \& \\ Vehicles\end{tabular}      & \begin{tabular}[c]{@{}c@{}}Common vehicles that appear on\\ work-site, like excavators.\end{tabular}                            & Forbidden    & 0.35\%       \\ \hline
Obstacles                                                             & \begin{tabular}[c]{@{}c@{}}Uncommon objects that need to be\\ avoided, like steel bar, sign block, etc.\end{tabular}                    & Forbidden    & 0.23\%       \\ \hline
\end{tabular}
}
\caption{\textbf{CWT Ontology:} Classification of terrain features used in our approach}
\label{tab:onto}
\vspace{-25pt}
\end{table}

The CWT dataset is annotated with seven labels based on terrain features and navigability, as shown in Table~\ref{tab:onto}.
The annotation is decided based on the opinion of a team of excavator operators. 
% Based on their experience, flat surfaces, bumpy surfaces, and mixtures of water and dirt are usually navigable. 
In most cases, when flat surfaces are detected, they are preferable to other surfaces. 
% Water is a forbidden region, since it is difficult to gauge how deep the water is, and the soil near the water could be soft and easily deformed. 

% Mark this part TBD

While the CWT dataset and other datasets like RUGD~\cite{RUGD2019IROS} and RELLIS-3D~\cite{rellis} are collected in unstructured, outdoor environments, the CWT has several distinctions. As shown in Figure~\ref{fig:dataset_pic}, the CWT dataset mostly consists of uneven terrain with unfavorable road conditions and covers many situations that might be encountered on a work site, including rock-piles, pits, stagnant water after rain, etc.

In addition, the CWT dataset focuses entirely on roads and terrains, and the annotation is based on terrain semantics instead of fine-grained semantics on every possible classes. Such annotation scheme is designed for the benefit of other downstream tasks, including planning and navigation for robots of any sizes, and excavation activities on hazardous terrains.

Overall, CWT presents many new challenges to the vision community to improve perception in hazardous environment, while providing support for autonomous robotics applications in dangerous environment. We demonstrate the difficulty of our dataset by showing the performances of several SOTA semantic segmentation methods on the CWT and existing off-road datasets like RELLIS-3D in Section~\ref{eval_data}. The CWT dataset can be accessed through \href{https://forms.gle/zeAcgptpideCrFbw8}{this link}.

% Nevertheless, we still give forbidden region a traversability score of 0.2 just in case the excavator has to navigate through this region to the goal. 
% Even though the excavator can usually traverse through shallow water, we choose to minimize the risk and damage. Obstacles like rocks and other excavators must be avoided for safety. 
% The annotation is decided and implemented based on the opinion of a team of excavator operators.
% This dataset will be released for future research in perception in off-road and unstructured environments.

\section{Experiments and Evaluations}
\label{evaluation}

In Section~\ref{eval_data}, we show evaluation results for the semantic segmentation task on our CWT dataset and RELLIS-3D~\cite{rellis}. In Section~\ref{eval_map}, we evaluate our \ours{} on RELLIS-3D and show the benefits of our method compared to other SOTA mapping methods.

\subsection{Perception Evaluation on the CWT Dataset}
\label{eval_data}

We show some evaluations using several SOTA segmentation methods on the CWT dataset and the RELLIS-3D dataset in Table~\ref{tab:flops}. The CWT dataset is a more challenging terrain dataset than RELLIS-3D. We also highlight the number of parameters and Giga-FLOPS (floating-point operations per second) as a measurement since energy efficiency is an important factor for robotic applications. The method and evaluation for segmentation is based on MMSeg~\cite{mmseg2020}.

% Please refer to the supplemental material for more detailed comparisons between our CWT dataset, the RUGD~\cite{RUGD2019IROS} dataset, and the RELLIS~\cite{rellis} dataset.

% \input{table/perception_table}

\subsection{Terrain Traversability Map Evaluation}
\label{eval_map}

In Table~\ref{tab:trav}, we evaluate the accuracy of our method and compare it with several SOTA traversability mapping methods on the RELLIS-3D dataset. We use the ground truth semantic labels from RELLIS-3D on a 3D point cloud and convert the labels to either 0 or 1 to indicate traversability on a grid map.
During evaluation, we assume that the traversability map is based on the Clearpath Warthog, the same robot that collected the RELLIS-3D dataset: traversable regions like grass, dirt, concrete, and asphalt are set to 0, while puddles, bushes, and obstacles are set to 1. Even though our method outputs a continuous value between 0 and 1, we want to simplify the conversion between labels and traversability scores to avoid any biases.
% \vspace{-10 pt}
\subsubsection{Comparisons}

Since many methods do not have publicly available codes, we implement their methods based on the papers, which can only run on an offline dataset and not in the real world. We compare our method with the following methods:

\noindent\textbf{Dahlkamp et al.~\cite{pc_desert_cmp1}} use a Mixture of Gaussian Model to make a binary prediction on RGB images for traversable regions and make an inverse perspective transform to the world coordinates.

\noindent\textbf{Sock et al.~\cite{3d_cam_cmp3}} use a Linear Support Vector Machine for a 2-classes prediction and some mapping between terrain slope and a traversability score between 0 and 1. The final map is obtained through Bayes Fusion of terrain classification and slope information.

% \noindent\textbf{Maturana et al.~\cite{semantic_only_cmp4} and Zhao et al.~\cite{semantic_for_planning_cmp2}} 
\noindent\textbf{Zhao et al.~\cite{semantic_for_planning_cmp2}} use a multi-class segmentation method based on RGB images and make projections onto a grid map for planning and navigation. Maturana et al.~\cite{semantic_only_cmp4} use a distance transformation and update new observations with Bayes's rule.

\noindent\textbf{Geometric-based methods~\cite{geo1, geo_trav}} only use geometric information from the point cloud for navigation tasks.

\noindent\textbf{3D semantic segmentation~\cite{kpconv,salsanet} methods} are useful for classifying terrains. We obtain their inference results from the official repository of RELLIS-3D~\cite{rellis}.

\begin{table}[t]
% \newcolumntype{Z}{S[table-format=2.2,table-auto-round]}
\centering
\vspace{7pt}
\Large
\begin{adjustbox}{max width=\columnwidth}
\begin{tabular}{@{} llccccc @{}}
  \toprule[2 pt]
  {Methods} & {Params $\downarrow$} & {Dataset}  & {mIoU $\uparrow$} & {mAcc $\uparrow$} & {Img Size} & {GFLOPs $\downarrow$}\\
  
  \midrule

    %  \multirow{2}[1]{*}{RUGD} 

\multirow{2}[1]{*}{CGNet~\cite{cgnet}} & \multirow{2}[1]{*}{\textbf{0.494 M}} & CWT & 53.41 & 67.59 & 1920 x 1080 & 27.62\\
& & RELLIS & 65.9 & 79.25 & 1920 x 1200 & 30.67\\
\midrule
\multirow{2}[1]{*}{Fast SCNN~\cite{fastscnn}} & \multirow{2}[1]{*}{1.45 M} & CWT & 54.77 & 68.75 & 1920 x 1080 & \textbf{7.45}\\
& & RELLIS & 69.27 & 80.99 & 1920 x 1200 & \textbf{8.03}\\
\midrule
\multirow{2}[1]{*}{Fast FCN~\cite{fastfcn}} & \multirow{2}[1]{*}{68.7 M} & CWT & 41.68 & 51.85 & 1920 x 1080 & 1031.51\\
& & RELLIS & 68.24 & 79.21 & 1920 x 1200 & 1145.6\\
\midrule
\multirow{2}[1]{*}{BiSeNetV2~\cite{bisenetv2}}  & \multirow{2}[1]{*}{14.77 M} & CWT & 54.37 & 67.05 & 1920 x 1080 & 97.51\\
& & RELLIS & 65.33 & 75.06 & 1920 x 1200 & 108.38\\
\midrule
\multirow{3}[1]{*}{SETR*~\cite{setr}} & \multirow{3}[1]{*}{109.67 M} & CWT & 19.91 & 30.61 & 1920 x 1080 & -- \\
& & RELLIS & 65.53 & 76.57 & 1920 x 1200 & --\\
& & -- & -- & -- & 1024 x 512 & 337.46$\dagger$ \\
\midrule
\multirow{3}[1]{*}{DPT*~\cite{dpt}} & \multirow{3}[1]{*}{309.17 M} & CWT & 29.02 & 47.65 & 1920 x 1080 & -- \\
& & RELLIS & 55.38 & 66.23 & 1920 x 1200 & --\\
& & -- & -- & -- & 1024 x 512 & 424.87$\dagger$\\
\midrule
\multirow{2}[1]{*}{Segformer~\cite{segformer}} & \multirow{2}[1]{*}{3.72 M} & CWT & 50.6 & 64.29  & 1920 x 1080 & 50.55$\dagger$\\
& & RELLIS & 68.62 & 83.4 & 1920 x 1200 & --\\
% & & -- & -- & -- & 1024 x 512 & 12.77$\dagger$\\

  \bottomrule[2 pt]
\end{tabular}

\end{adjustbox}
\caption{\textbf{Perception Accuracy on the CWT and RELLIS-3D~\cite{rellis} Dataset:} We list several SOTA semantic segmentation methods and train the model with 240K iterations. The CWT dataset has lower accuracy compared to the RELLIS dataset. $*$ marks methods that do not converge well after 240K additional iterations. $\dagger$ marks the GFLOPs as an approximation and a lower bound.}
\vspace{-7mm}
\label{tab:flops}
\end{table}

\begin{table*}[t]
% \newcolumntype{Z}{S[table-format=2.2,table-auto-round]}
\centering
\vspace{7pt}
\Large
\begin{adjustbox}{max width=\textwidth}
\begin{tabular}{@{} lccccccc @{}}
  \toprule[2 pt]
{Methods} & {Modality} & {Goal} & {Trav / Non-Trav Acc $\uparrow$} & {mAcc $\uparrow$} & {aAcc $\uparrow$} & {AUC $\uparrow$} & {MSE $\downarrow$}\\
  
  \midrule

    %  \multirow{2}[1]{*}{RUGD} 

KPConv*~\cite{kpconv}  & LiDAR & 3D segmentation & 33.33 / 79.24 & 56.28 & 67.65 & - & 0.253 \\
SalsaNet*~\cite{salsanet} & LiDAR & 3D segmentation & 94.82 / 57.75 & 76.28 & 67.11 & - & 0.370\\

  \midrule
  \midrule

Chilian et al.~\cite{geo1} & LiDAR & Mapping \& Navigation & 66.19 / 88.29 & 77.24 & 82.17 & 0.790 & 0.155\\

Dahlkamp et al.~\cite{pc_desert_cmp1} & RGB Camera & Navigation & 4.41 / 99.91 & 52.16 & 54.42 & 0.751 & 0.123\\
Zhao et al.~\cite{semantic_for_planning_cmp2} & LiDAR + Stereo Camera & Mapping \& Navigation & 9.31 / 99.85 & 54.58 & 56.67 & 0.528 & 0.128\\
Sock et al.~\cite{3d_cam_cmp3} & LiDAR + RGB Camera & Mapping \& Navigation & 1.93 / 99.93 & 50.93 & 53.21 & 0.590 & 0.156\\
\ours{} (ours) & LiDAR + RGB Camera & Mapping \& Navigation & 71.77 / 91.05 & \textbf{81.41} & \textbf{85.70} & \textbf{0.803} & \textbf{0.106}\\

  \bottomrule[2 pt]
\end{tabular}

\end{adjustbox}
\caption{\textbf{SOTA comparisons:} We list several prior methods and highlight the benefits of our method on the RELLIS~\cite{rellis} terrain dataset. Our method outperforms previous SOTA methods by 4.17-30.48\% in terms of mAcc and reduces the MSE by 13.8-71.4\%.}
\vspace{-4.5mm}
\label{tab:trav}
\end{table*}

\begin{figure*}[t!]
    \centering
    \includegraphics[width=0.9\textwidth]{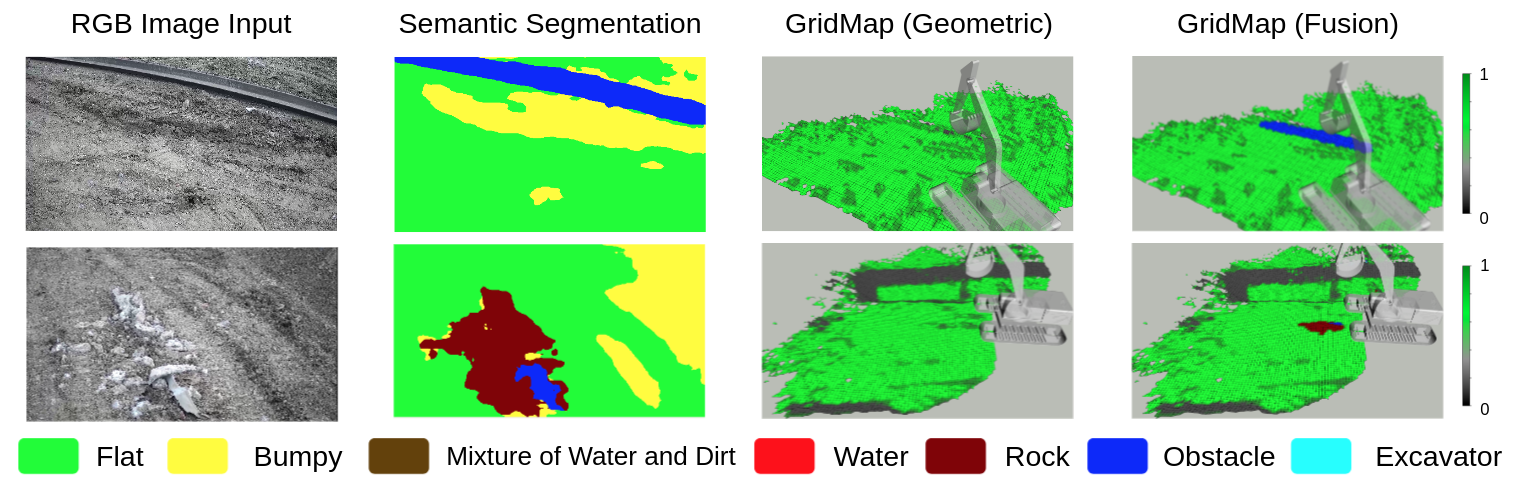}
    \caption{\textbf{Visual results of \ours:} 
    % Each row shows the input RGB image, the corresponding semantic segmentation result, the traversability map with only geometric information, and the traversability map after fusion. 
    In the traversability map, the higher the traversability score, the easier it is for robots to navigate the corresponding terrain. 
    % We use the color scheme from green to grey to highlight some semantic labels like rock, obstacle and water. 
    % We see that there are many places that are not detected through their geometric properties, but that can be recognized using their visual features; those are reflected in the final traversability map.
    More visual results are available in the supplementary material.}
    \label{fig:output}
    \vspace{-20pt}
\end{figure*}

\subsubsection{Evaluation Metrics and Results}

We evaluate the traversability map based on offline data with four different metrics. In general, our method has better performance in terms of accuracy and MSE. Note that in the first three metrics, all traversability values are converted to either 0 or 1 for methods that have a continuous output. The metrics are described as follows:

\noindent\textbf{Mean Accuracy:} The average accuracy of traversable and non-traversable regions.

\noindent\textbf{All Accuracy:} Accuracy over all grids.

\begin{figure}[t]
    \centering
    \includegraphics[width=0.9\columnwidth]{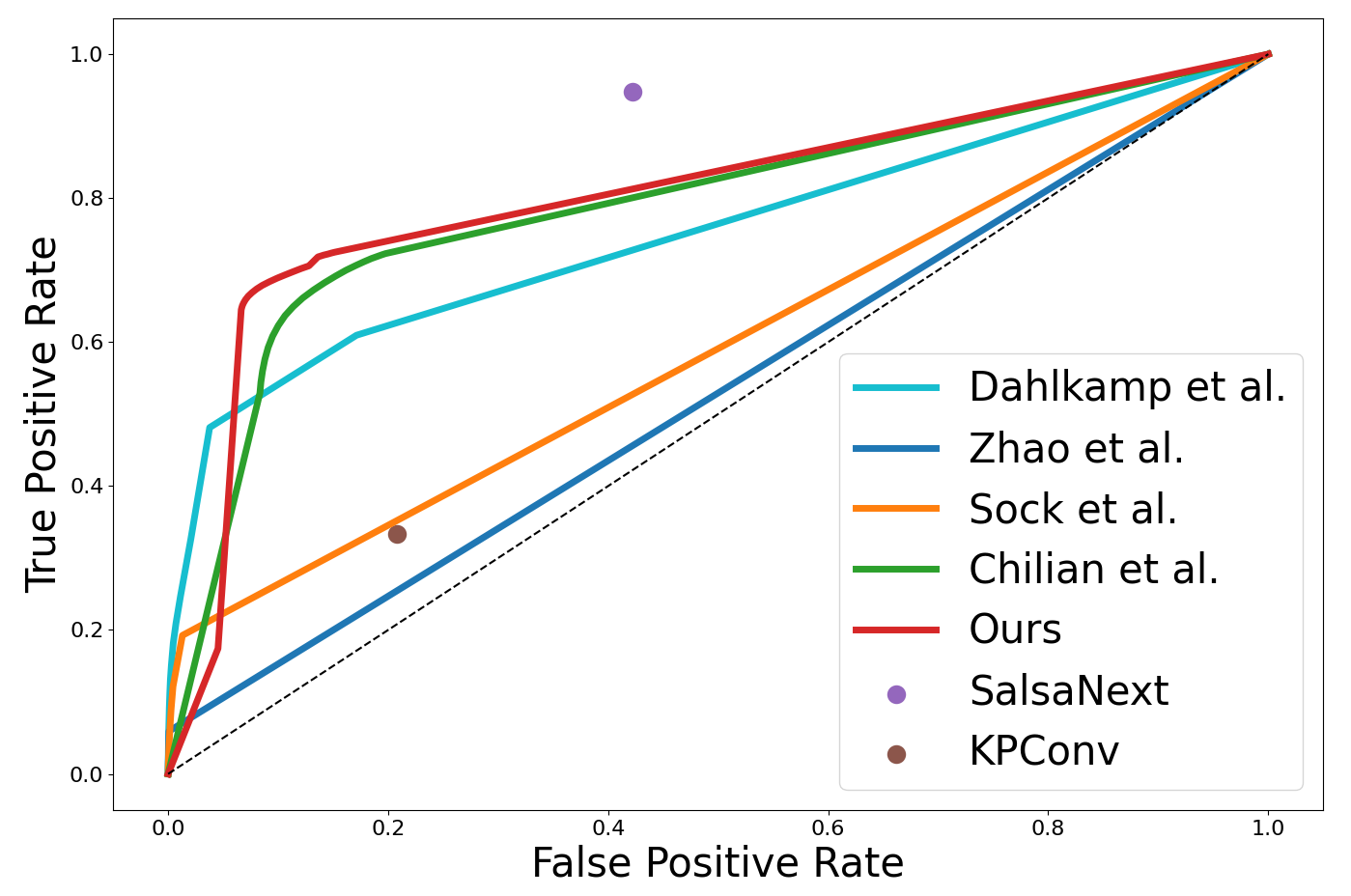}
    \caption{\textbf{ROC plot:} We plot ROCs on several SOTA mapping/segmentation methods. LiDAR-based segmentation methods~\cite{salsanet, kpconv} are trained on the point cloud labels, so they have the advantage of prior knowledge on the ground truth. In the real world, annotated 3D point cloud data would not be easily available for applications. We use a point in the ROC plot to represent those methods, as there is not a threshold to adjust.}
    \label{fig:roc}
    \vspace{-14pt}
\end{figure}

\noindent\textbf{ROC (Receiver Operation Curve):} Previous methods~\cite{pc_desert_cmp1, 3d_cam_cmp3} make binary predictions over each grid, so ROC is a common indicator of the performance through true positive and false positive rates, as shown in Figure~\ref{fig:roc}.

\noindent\textbf{MSE (Mean Squared Error):} To describe how well the prediction fits the ground truth, we also calculate the average distance between the prediction and the ground truth over all grids.

\vspace{-7pt}
\section{Performance in Real-World Environments}
\label{aes_imp}

In this section, we highlight the results on real-world environments and overall performance of our navigation system based on \ours{}. We also compare its performance with a geometric-only method~\cite{geo1}. 
% This method is referred to as the geometric-only method~\cite{geo1} in the rest of this section.

% \subsection{Implementation Details}
\subsection{Hardware Setup}
\label{hardware}
We use an XCMG XE490D excavator to perform our experiments. The excavator is equipped with a Livox-Mid100 LiDAR, an HIK web camera with FOV of 56.8 degrees with a pitch angle of 30.3 degrees to detect the environment, and a Huace real-time kinematic (RTK) positioning device to provide the location. We run our code on a laptop with an Intel Core i7-10875H CPU, 16 GB RAM, and 6GB GeForce RTX 2060 on the excavator. 

XCMG XE490D excavator has a maximum climbing angle of 35 degrees; the typical recommended climbing angles for any vehicle as a safe climbing angle is 10 degrees. Therefore, we set $s_{cri} = 35\ deg$ and $s_{safe} = 10\ deg$. In addition, we obtain an approximation of the maximum height allowed by $s_{cri}$ and $s_{safe}$ after expanding three times the resolution $d_{res}$ along the surface to get:
\begin{align*}
h_{cri} &= 3\ tan(s_{cri}) \times d_{res} = 0.35\ m\\
h_{safe} &= 3\ tan(s_{safe}) \times d_{res} = 0.10\ m
\end{align*}

\begin{figure}[t]
% 	\vspace{5pt}
    \centering
    \includegraphics[width=\columnwidth]{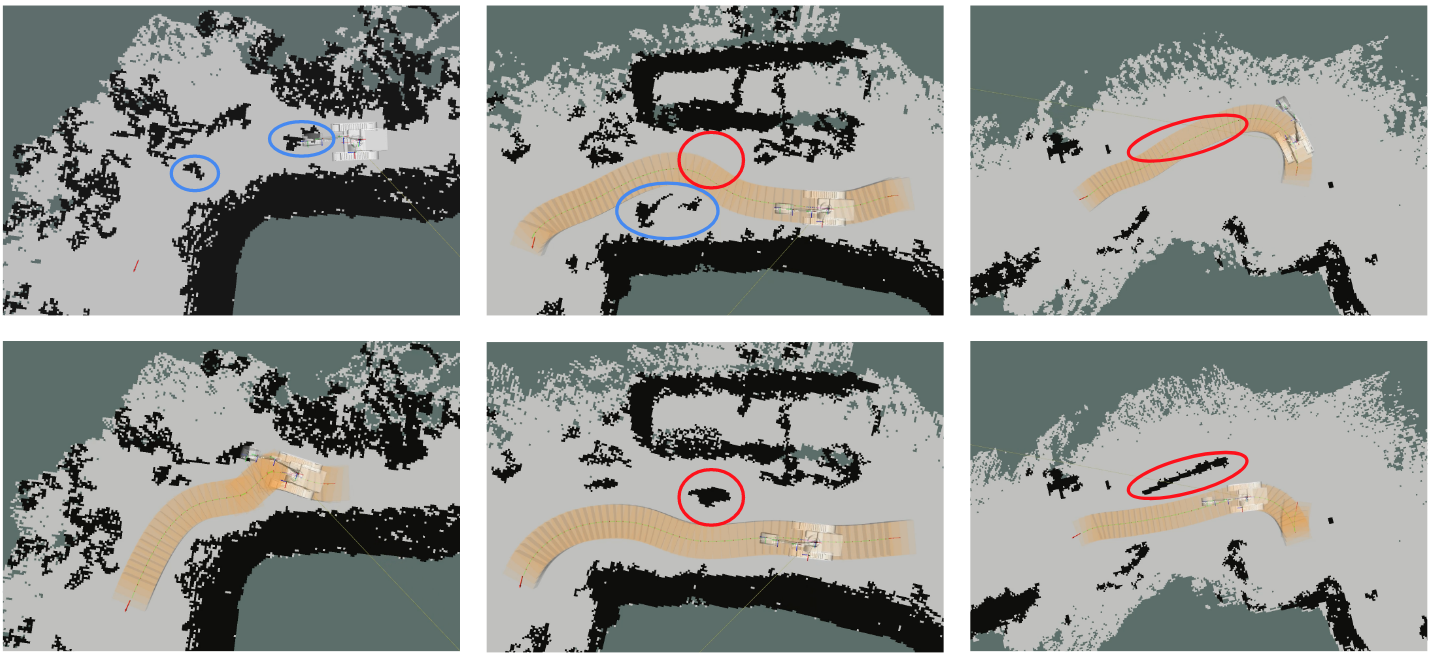}
	\vspace{-10pt}
    \caption{ \textbf{Planner output comparisons between geometric-only scheme~\cite{geo1} (top) and \ours{} (bottom):} We show planned trajectories with our modified Hybrid A*~\cite{hybridA} planner. The planning is based on a global traversability map. We highlight some obstacles that are not observed by the geometric method (\textcolor{red}{red}) as well as some traversable regions that are falsely observed by the geometric method (\textcolor{blue}{blue}). }
	\label{fig:planner}
	\vspace{-10pt}
\end{figure} 

\vspace{-5pt}

\subsection{Traversability Map Results and Analysis}

In this section, we evaluate our system in the real world with visual results. In Figure~\ref{fig:output}, we show some typical scenarios excavators encounter to illustrate the advantages of geometric and semantic fusion. 
% For visualization purposes, we change the color scheme to range from green to grey and paint the detected obstacles and forbidden regions like water to the corresponding color in Figure~\ref{fig:output}. 
In those cases, the steel bar and stone were not captured by geometric calculation, while with semantic information, those obstacles can be detected.

\subsection{Planning Based on Offline Traversability Map}

Based on the resulting occupancy grid maps from the proposed \ours{} and geometric-only method~\cite{geo1}, we randomly choose start and goal positions on an unoccupied grid with over 90 trials. The success rates of finding a valid path without collision for our \ours{} and the other method are 82.6\% and 33.3\%, respectively.
We show some comparisons on planning results in Figure~\ref{fig:planner}. 
% The implementation is based on publicly released code. 
We use an occupied threshold $t_{occ}$ of 0.6. The height of the cabin $h_{cab}$ is $0.5\ m$, and the distance between two tracks $d_{track}$ is $2.75\ m$ for map post-processing and planner configuration. 
% Please refer to the supplemental material for more experiment details.

\subsection{Real-world Experiments and Trials}

We test our system \ours{} on two construction sites with a total area of at least 200 $m^2$. We summarize those trials in Table~\ref{tab:trials}. We tested 3 types of trajectories, including going straight while avoiding lower traversability areas, making normal turns, and making sharp turns on the terrain. For all tests, the excavator was able to successfully reach the given target, which demonstrates the robustness of our system. Furthermore, the tracking error of all trajectories is within 10cm on average.
For details of the testing site, please refer to the supplemental materials.
% and log of each trials

\begin{table}[t]
\centering
\resizebox{1\columnwidth}{!}{%
\begin{tabular}{lcccc}
  \toprule
Trajs & Type  & Total Len (m)  & Avg Err (m) &Min Err (m) \\ 
  \midrule
9            & Straight    & 158.85 & 0.102    & 0.040 \\
8            & Small Turn    & 219.35  & 0.104 & 0.032\\
8            & Sharp Turn   & 242.14   & 0.059 & 0.042\\ 
% T-Calculation(average)            & - & -  & - & 60.8\\ 
%   \midrule
% Overall            & \he{0} & \he{0} & \he{0} & \he{0} & \he{0} \\
\bottomrule
\end{tabular}
}
\caption{\textbf{Real-world Experiments and Trials.} we have tested \ours{} on different types of trajectories, including straight paths, normal turns, and sharp turns. Our system can achieve $10$cm tracking error accuracy for all these scenarios.}
\label{tab:trials}
\vspace{-18pt}
\end{table}

% # of Trj	Type	Total Length (m)	Avg Error (m)	Min Error (m)	Max Error (m)
% 9	Straight	158.849	0.102	0.040	0.287
% 8	Small Turn	219.348	0.104	0.032	0.253
% 8	Sharp Turn	242.149	0.059	0.042	0.143

% In the last case, the rough and bumpy region is in fact water and should not be traversed.

% In Figure~\ref{fig:qualitative}, we compare traversability maps generated using a geometric-only method~\cite{geo1} and using TTM with geometric-semantic fusion. The output after fusion is less noisy since segmentation results can smooth out safe regions. Our method detects more non-traversable regions based on obstacles and dangerous regions from semantic information. 
% % For more comparisons, please refer to the supplemental material.

% \begin{figure}[t]
%     \centering
%     \includegraphics[width=\columnwidth]{fig/occ4.png}
%     \caption{\textbf{Gridmap comparison between the geometric-only scheme~\cite{geo1} and ours:} (1) Our method is less noisy and has more connected regions to plan a feasible trajectory. (2) Our method can detect obstacles that the geometric method could not recognize.}
%     \label{fig:qualitative}
%     \vspace{-10pt}
% \end{figure}

\vspace{-5pt}
\subsection{Run-time Analysis of Traversability Map}
Our method consists of the following major parts, which contribute to the overall runtime of the system:

\begin{figure}[t]
    \centering
    \includegraphics[width=\columnwidth]{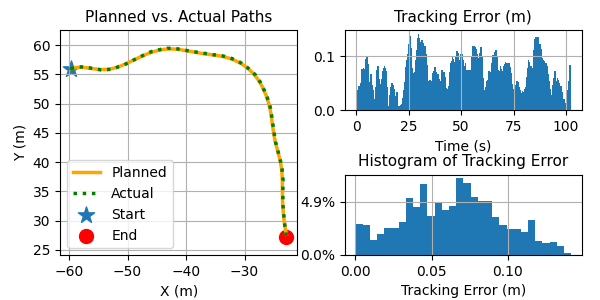}
    \vspace{-8pt}
    \caption{\textbf{Tracking controller performance evaluation:} We plot one of the trajectory tracking results with planned and actual paths (\textbf{left}), tracking error vs. time (\textbf{top-right}), and the histogram of the tracking error (\textbf{bottom-right}).}
    \label{fig:track_err}
    % \vspace{-16pt}
    \vspace{-10pt}
\end{figure}

\begin{table}[t]
\centering
\resizebox{0.7\columnwidth}{!}{%
\begin{tabular}{lccc}
  \toprule
Run-time (ms) & Max  & Min  & Mean  \\ 
  \midrule
Segmentation            & 100.2    & 53.5 & 75.4\\
Projection                & 54.0    & 35.0 & 42.3\\
$T_{geo}$ Calculation            & 38.0  & 9.0 & 22.1\\ 
% T-Calculation(average)            & - & -  & - & 60.8\\ 
%   \midrule
% Overall            & \he{0} & \he{0} & \he{0} & \he{0} & \he{0} \\
\bottomrule
\end{tabular}
}
\caption{\textbf{Runtimes of different modules.} Our method can be run in real-time and update the traversability map at a rate of 10 Hz. }
\label{tab:runtime}
\vspace{-20pt}
\end{table}

% \vspace{-6pt}
\begin{itemize}
    \item \textbf{Segmentation} generates a pixel-wise semantic classification on each image in the RGB input stream.
    \item \textbf{Projection} casts the 2D segmentation result onto the 3D point cloud and assigns each point a semantic label through the calibration matrix. 
    \item \textbf{Geometric traversability calculation} estimates and updates slope and step height based on point cloud data in a grid map representation.
    % It constructs a map through raw point clouds, calculates the geometric information, and then fuses geometric information with semantic information to calculate Traversability score $T$.
\end{itemize}
% \vspace{-7pt}

% \tian{Claim here that fusion step is very fast (under xxx ms).}
% It should be noted that the algorithm does not update the geometry information for each LiDAR scan but only performs one calculation after 3 LIDAR scans are input because the excavator moves slowly, and most of the areas between two consecutive frames overlap. In this way, we can save computing resources and boost the run-time of the overall system.

% As shown in Table~\ref{tab:runtime}, the running time per update of each component mentioned above is 75.4 ms, 42.3 ms, and 22.1 ms, respectively. 

% Since we calculate $T_{geo}$ for every three point cloud inputs, the run-time of the geometric traversability calculation is 60.8 ms per input.

In Table~\ref{tab:runtime}, we give details of the run-time of each component in the system.
The final fusion step is under 2 ms and does not contribute to the overall runtime of the method.
% Our method can handle an RGB image stream of 25 Hz and a point cloud stream of 10 Hz in the real world without lagging, and the map can update semantic and geometric information at a rate of 10 Hz. 
Our method can update the traversability map at a rate of 10 Hz.
Please refer to the video for more visual results of excavator navigation.

\subsection{Controller Error Analysis}

The tracking trajectory controller can maintain the excavator around the desired path with the maximum absolute lateral tracking error less than 15 cm in most of our test runs. In the case shown in Figure~\ref{fig:track_err}, the maximum tracking error is around 14 cm. This test run lasts for 102 seconds with an average speed of 0.5 meters per second. The total length of the trajectory is about 50 meters. The left plot shows the planned path from the improved hybrid A* planner and the actual path of the excavator. The excavator starts from the blue pentagram and ends at the red dot. The top-right plot represents the tracking error, which is the distance between the excavator and the closest point on the planned path. The bottom-right plot is the histogram of the tracking error, where the y-axis represents the percentage of each tracking error column. The tracking error is around 6 cm most of the time.

\subsection{Analysis and Lessons Learned}

In this section, we highlight some of the failures of and lessons learned from the design and evaluation of our system:   
\begin{itemize}
    \item \textbf{Perception errors} include segmentation error and Lidar measurement error. It is hard to find similar terrains or scenarios in existing datasets for annotations and supervised training, especially when the terrain becomes rougher and bumpier. To alleviate segmentation error, we collect and annotate some terrain data on construction sites with different terrain labels, including flat surface, bumpy surface, water puddle, obstacles, rocks, etc., aiming to improve perception accuracy in unstructured environments and enable such construction vehicle applications. However, the Lidar measurement can be unreliable due to the dust in the air. To remove such noise, we use step height estimation and semantic fusion for more robust traversability predictions, as mentioned in Section~\ref{trav_mapping_A}.
    \item \textbf{Terrain roughness} is an issue in geometric-based traversability methods on a mobile robot~\cite{geo1}. However, it is less effective for large machines like excavators due to the scale difference. In our case, roughness can be partially modeled either through the slope and step height or captured by visual features from the RGB images. However, it could become an issue if the terrain is very uneven or has large rocks or obstacles.
    \item \textbf{Localization accuracy} directly impacts the quality of the system. % including aspects such as the accuracy of the angle and height. 
    In our experiment, the main reason for the localization inaccuracy is the drift of the RTK system on the altitude. 
    In our open test field, the accuracy of the RTK system in latitude and longitude is around 5 cm, whereas the altitude accuracy is about 20 cm.
    To plan accurately and navigate over a period of time, we only use the most recent grid cells to calculate the traversability score because the drift is small. 
    % We will also adjust the drift accordingly so that the latitude and longitude is still accurate.
    In addition, our attempt to use SLAM for localization failed because most features are quite uniform (similar hills, pits, rock piles, etc.), causing degraded performance and very low accuracy due to instability. In the future, we could build a more stable localization system to fuse RTK, LiDAR, and camera data. 
%     we also try to combine Lidar SLAM which can maintain high accuracy for robot localization in our work. 
% However, due to the small field of view of the lidar we use, the lack of features of the field, resulting in SLAM pose estimation degradation occurring from time to time, which will cause the map split. In the future, we will build a more stable localization system to fuse the data of RTK, Lidar, and camera. 
    % \item \textbf{Extensive evaluation} of such a system in different types of terrains and weather conditions is very important. The performance of the perception and traversability map computation system varies in different conditions, and that impacts the path computation and overall navigation.
    % We have evaluated our system over 30 different terrains, and plan to test it in more complex scenes. 
    \item \textbf{Planner} needs to be adjusted to fully utilize the traversability map. We choose the Hybrid A* algorithm over the standard A* algorithm in our system to avoid sharp turns, which could cause damage or bumpiness to the ground surface. We adjust the Hybrid A* planner as described in~\ref{planning} to compute a smoother and safer path with continuous traversability map values. However, it is hard to guarantee that our planner will always generate a smooth path on arbitrary terrains.
    \item \textbf{Computational and power budget} is a major issue in the design of our perception and planning algorithm. Our traversability map computations and navigation module run on a laptop with an Intel Core i7-10875H CPU and a 6GB GeForce RTX 2060. Our implementation must be efficient and light-weight to run in real-time. Recently, many deep and reinforcement learning methods have been proposed for object detection and navigation, but they require a high-end GPU for efficient execution. We can't use such methods on our platform. 
    % \item \textbf{Haptic feedback} can be very useful for navigation and excavation operations. Our current system only uses visual sensors in the form of LiDAR and cameras, but no force feedback. A multimodal interface with force and visual feedback can definitely improves its performance in terms of terrain navigation as well as overall task performance. 
    \item \textbf{Safety:} In deploying the autonomous excavator system to the real world, safety is always the most critical consideration. We develop the terrain traversability mapping component to describe the complexity of the terrain and provide safe regions for the autonomous excavator to navigate. Our method can be combined with other safety strategies such as object detection, collision avoidance, etc., and maintain the stability of the excavator to ensure the safety of autonomous operation.
    \item \textbf{Excavator size} also governs the performance of our system. There are three broad classes of excavators: compact excavator (less than 6 tons), standard excavator ($7-45$ tons) and large excavators ($45-90$ tons). The size of the excavator impacts the performance of the navigation system when computing a smooth trajectory and the resulting path. There is a relative trade-off between mobility and stability for different sizes. We have evaluated the performance of \ours{} on a large, $49$-ton excavator. In general, developing autonomous excavation technology for larger excavators is more challenging.
\end{itemize}

% We include more details in the supplemental material.
\vspace{-5pt}
\section{Conclusions, Limitations, and Future Work}

In this paper, we present a terrain traversability mapping and navigation System (\ours{}) for autonomous excavation navigation. We highlight its application and benefits on difficult excavator navigation tasks in real-world scenarios. We use a novel learning-based geometric fusion solution and demonstrate its benefits over prior mapping algorithms.  We also release the CWT dataset with challenging real-world scenes in unstructured construction sites for perception tasks. 
%We show a proof a concept and demonstrate a baseline system that can navigate heavy-machines through challenging terrains.

Our work has some limitations. Due to safety issues, we are not able to extensively test our system in all types of scenarios, including cases with many human workers and other machines. We have only evaluated the performance on a large, $49$-ton excavator.
% (especially in some dangerous scenarios where most regions are bumpy or steep) to avoid failure cases like flipping over and injuring of the operator.
% In the future, after capturing more navigation results, we hope to measure traversability with additional navigation metrics like success rate, time-to-goal, etc.
% Second, we have tested our performance on existing planners, which are not able to exploit the full benefits of traversability. For example, sometimes the excavator would have been able to run over small obstacles with the space between two tracks.
As part of our future work, we would like to improve the planner further and utilize the specifications of the excavator like a human operator. 
For example, the excavator should be able to run over small obstacles using the space between two tracks.
In addition, we would like to evaluate the performance in different types of outdoor terrains. Our longer-term goal is to enable autonomy and collaborations among machines or with humans on construction sites. This requires several systems and modules working together, including autonomous excavation, autonomous navigation, and human machine interactions.
% Our final goal is using this framework to guide the excavator to navigate on its own in extreme and hazardous environments.

%%%%%%%%%%%%%%%%%%%%%

\section*{ACKNOWLEDGEMENT}
% \noindent\textbf{Acknowledgement.} 
This work was done as as summer intern at Baidu RAL. We appreciate the discussion and support from Baidu RAL team.

%%%%%%%%%%%%%%%%%%%%%

\section*{APPENDIX}

\section{More Details of the CWT dataset}

Our dataset is collected at a construction site while an excavator is navigating through the work area. We collect 3 videos that total approximately 30 minutes; 669 images of size $1920\times1080$ with pixel-wise annotation are included in our dataset. Please refer to \href{https://forms.gle/zeAcgptpideCrFbw8}{this link} for the access to the CWT dataset.

% The giant robot in our task needs to recognize dynamic entities, avoid obstacles (such as stones, steel bars, etc.), and turn itself to the ideal terrain (flat and dry ground). Considering these requirements, the dataset annotation is implemented to reflect terrain traversability based on the opinion of a team of excavator operators. 

% The CWT dataset is annotated with seven labels. Flat surfaces, bumpy surfaces, and mixtures of water and dirt are usually navigable. In most cases, when flat surfaces are detected, they are preferable to other surfaces. Water is a forbidden region, since it is difficult to gauge how deep the water is and the soil near the water could be soft and easily deformed.  

\begin{figure*}[t]
    \setlength{\linewidth}{0.95\textwidth}
    \setlength{\hsize}{0.95\textwidth}
	\centering
	\subfloat{
		\includegraphics[width = 0.34\linewidth]{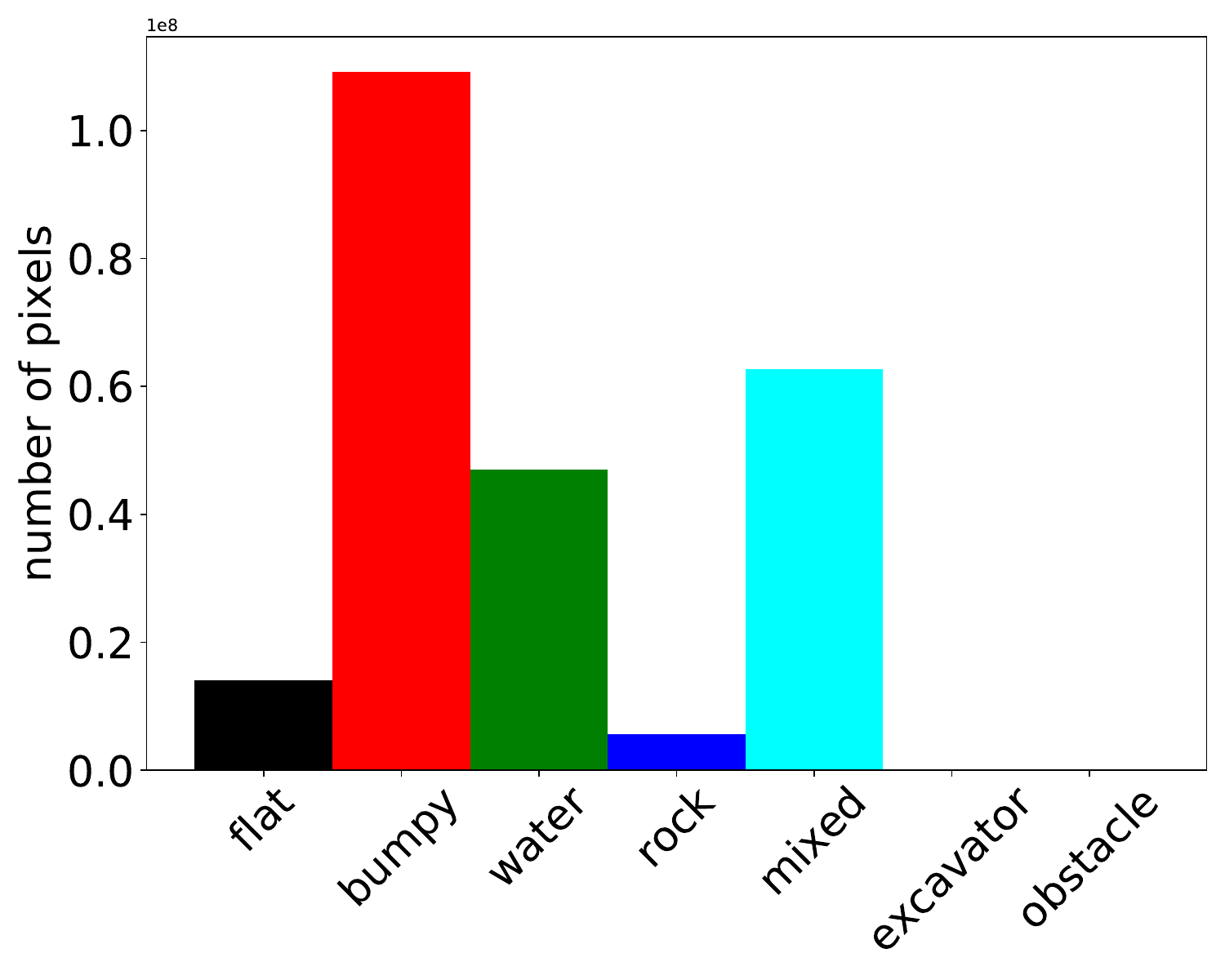} 
		\includegraphics[width = 0.34\linewidth]{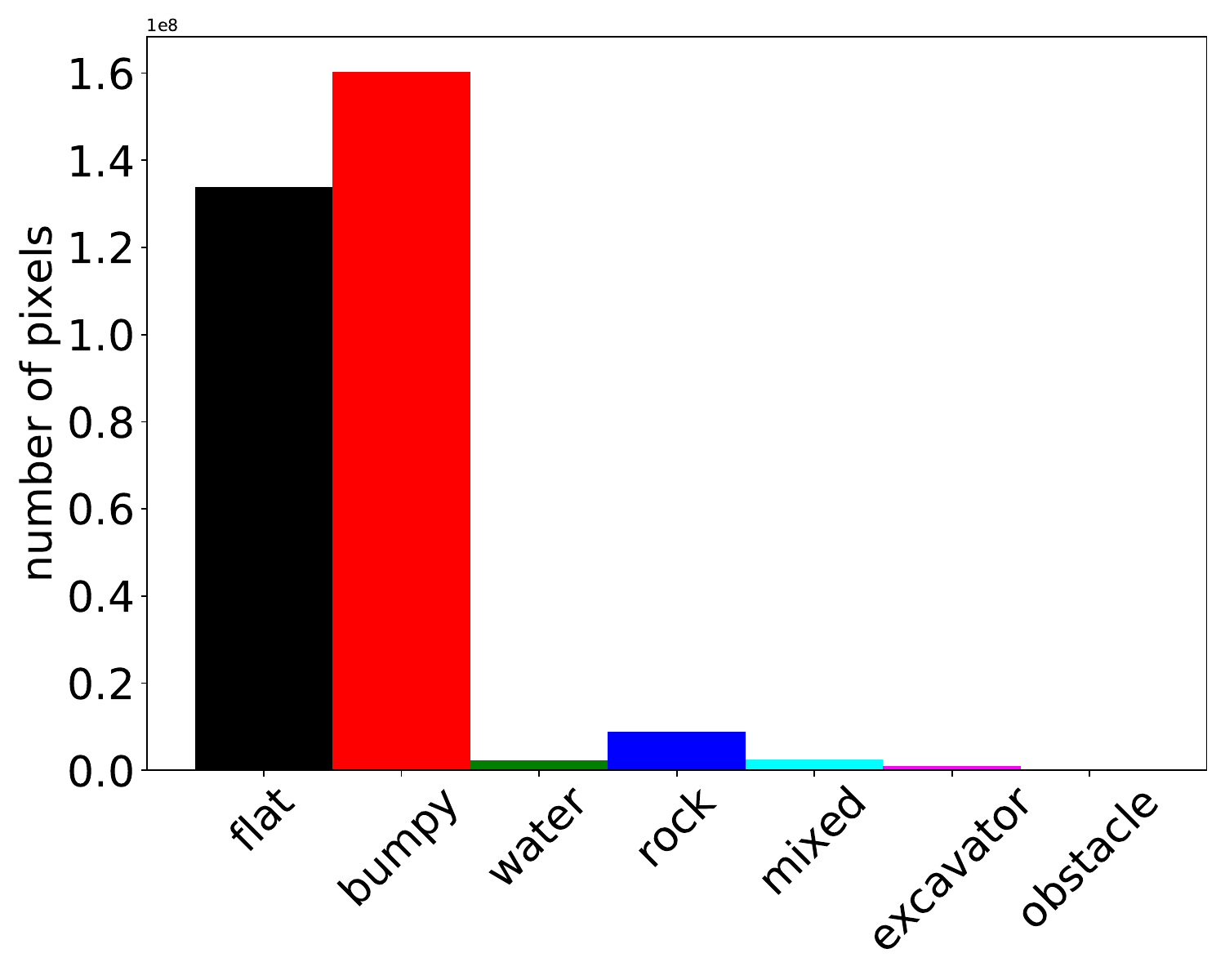} 
		\includegraphics[width = 0.34\linewidth]{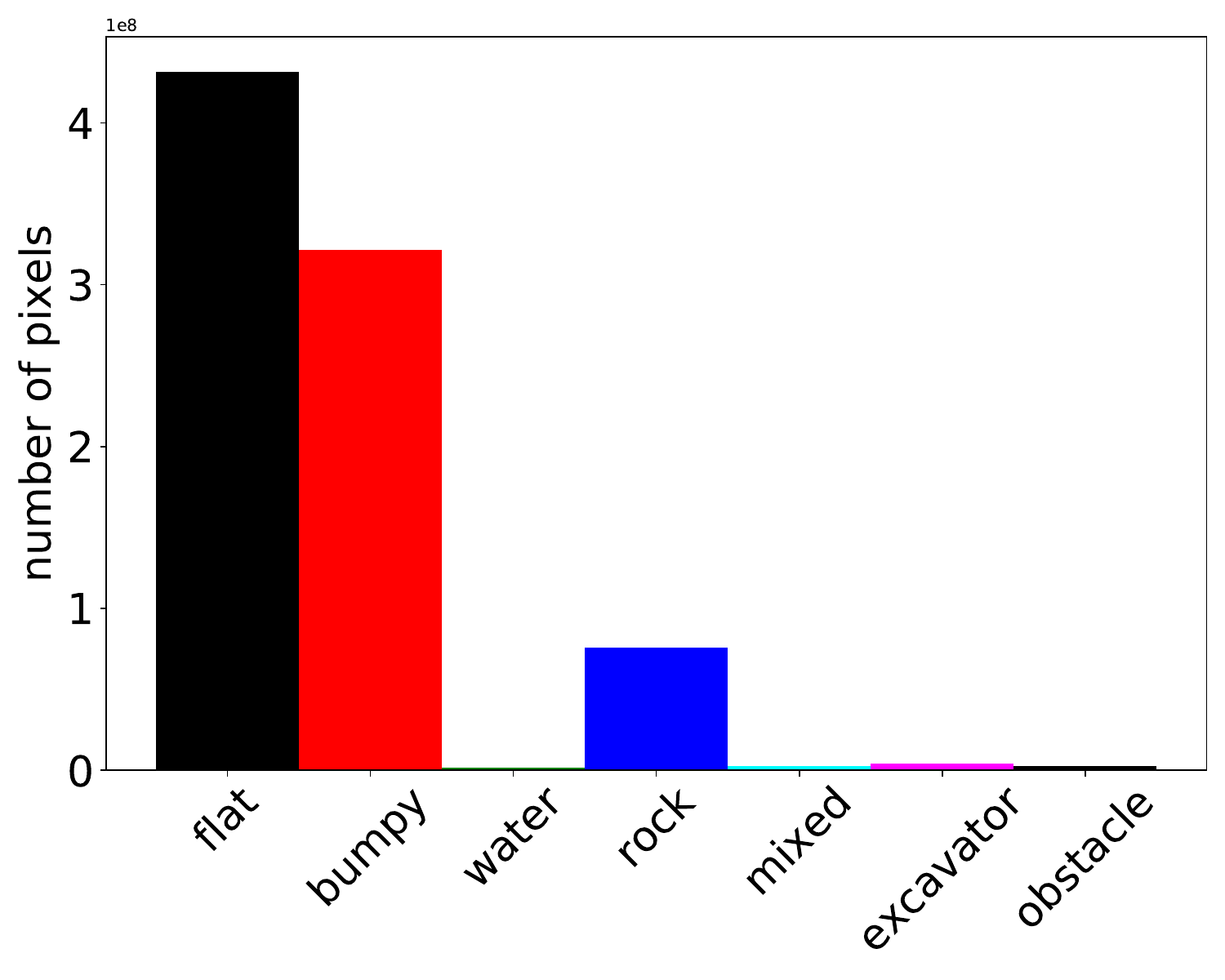} 
	} 		 
	\vspace{-3mm}
	\caption{\textbf{Label distribution of each image sequence.}}
	\vspace{-3mm}
	\label{fig:statistics}
\end{figure*}

\subsection{Dataset Details and Statistics}

There are three video sequences collected on our excavator test site. In Figure~\ref{fig:statistics}, we show the class distribution breakdown for three sequences. The first video is collected after rain and consists of mostly water and muddy ground. The trenches caused by excavation and navigation can also be seen. The other two video sequences are captured on a sunny day in different scenarios. The videos are collected by a professional operator controlling the movement of the robot. The three videos are 268s, 668s, and 822s. We sample the camera stream every two seconds and annotate the images with ground truth labels, resulting in a total of 669 images after removing some redundant ones.

\subsection{Benchmarks}
\begin{table*}[t]
% \newcolumntype{Z}{S[table-format=2.2,table-auto-round]}
\centering
% \setlength{\tabcolsep}{3mm}
% \Large
\begin{adjustbox}{max width=\textwidth}
\begin{tabular}{ lc|ccccccc|cc }
  \toprule[1 pt]
  {Year} & {Methods} & {Flat} & {Bumpy} & {Water} & {Rock} & {Mixed} & {Excavator} & {Obstacle} & {mIoU} & {mAcc} \\
  
  \midrule

    2018 & CGNet~\cite{cgnet} &  73.02 & 63.11 & 38.22 & 69.67 & 47.0 & 47.04 & 35.78 & 53.41 & 67.59\\
    2019 & Fast SCNN~\cite{fastscnn} & 74.1 & 65.87 & 32.02 & 73.42 & 46.58 & 45.51 & 45.91 & 54.77 & 68.75 \\
    2019 & Fast FCN~\cite{fastfcn} & 71.96 & 61.23 & 35.61 & 60.06 & 35.3 & 0.0  & 27.6 & 41.68 & 51.85\\
     2021 & BiSeNetV2~\cite{bisenetv2} &  76.49 & 69.65 & 38.33 & 71.44 & 46.42 & 41.02 & 37.22 & 54.37 & 67.05 \\
    2021 & SETR*~\cite{setr} & 54.24 & 49.67 & 4.07 & 25.23 & 6.03 & 0.0 & 0.16 & 19.91 & 30.61 \\
    2021 & DPT*~\cite{dpt} & 59.45 & 53.75 & 23.78 & 33.69 & 26.0 & 0.0 & 6.49 & 29.02 & 47.65\\
    2021 & Segformer~\cite{segformer} & 73.44 & 64.47 & 39.62 & 70.29 & 43.81 & 30.48 & 32.07 & 50.6 & 64.29   \\

  \bottomrule[1 pt]
\end{tabular}

\end{adjustbox}
\caption{\textbf{Performance of SOTA methods on the CWT dataset:} We list several SOTA semantic segmentation methods and train the model with 240K iterations. $*$ marks methods that do not converge well after 240K additional iterations.}
\label{tab:perception}
\vspace{-4mm}
\end{table*}

We give several metrics and show the performance of several SOTA methods on the CWT dataset in Table~\ref{tab:perception}. $C$ denotes the set of all classes.

$$
mIoU = 1/c \sum_{c=1}^{C} \frac{TP_c}{TP_c + FP_c + FN_c}
$$

$$
mAcc = 1/c \sum_{c=1}^{C} \frac{TP_c}{TP_c + FN_c}
$$

$$
aAcc =  \frac{\sum_{c=1}^{C}\ TP_c}{Numbers\ of\ All\ Pixels}
$$

\section{More Details of TNS}

\subsection{Roughness in Geometric Traversability}
\label{roughness}

\begin{figure}[t]
    \centering
    \includegraphics[width=\columnwidth]{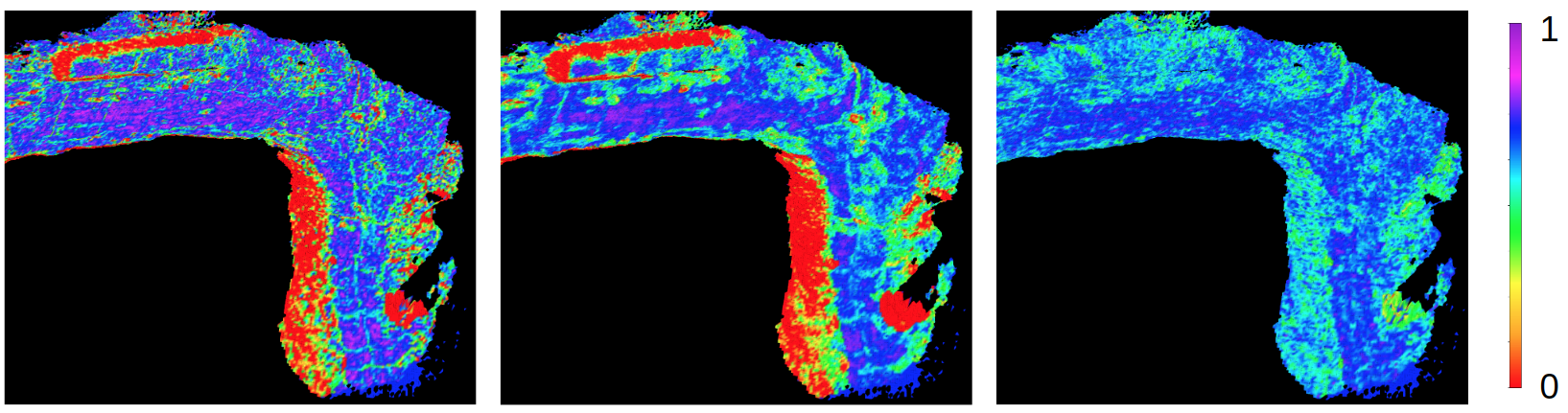}
    \caption{\textbf{Grid map output comparison:} We show slope \textbf{(left)}, step height \textbf{(middle)}, and roughness \textbf{(right)} values in geometric traversability computations. All values are converted to the scale from 0 to 1. Slope value tends to have many small areas of peaks, while step height tends to have smoother values across a bigger region. On the other hand, roughness does not have many peak values, and many regions like hills are already captured by the previous two measurements.}
    \label{fig:roughness}
    \vspace{-5mm}
\end{figure}

In many existing works~\cite{geo1, geo2} for geometric traversability or danger value calculation, a roughness score is calculated as a factor of terrain traversability.

\noindent\textbf{Roughness Estimation:}
The terrain roughness $r$ is calculated as the standard deviation of the terrain height values to the fitting plane. The distance $d$ from the center point $p$ of the grid to the fitting plane of $k$ neighboring grids is calculated as:
$$
d = \frac{\vec{p\bar{p}} \cdot \vec{n}}{| \vec{n} |}
$$
where $\vec{n}$ is the surface normal vector of the fitting plane and $\bar{p}$ is a point in the plane.
Finally, the roughness estimation of the grid $g$ can be computed as:
$$
r = \sqrt{ \sum_{i=1}^{k} (d_i)^2 }
$$

However, roughness is not a good measurement in unstructured environments, especially in our situation. During the design of our method, we discover that the roughness measurement is either random or, in some regions, the distribution of roughness resembles that of slope or step height, except with lower peak values. Eventually, the roughness score did not impact the results too much. 
We demonstrate such similarity in Figure~\ref{fig:roughness}.

\section{More Details of TNS-based Planning}

In this section, we add more details of our planning method and experimentation. 

\subsection{A* and Hybrid A*}

\begin{figure}[t]
	\centering
	\subfloat{
		\includegraphics[width = 0.45\linewidth]{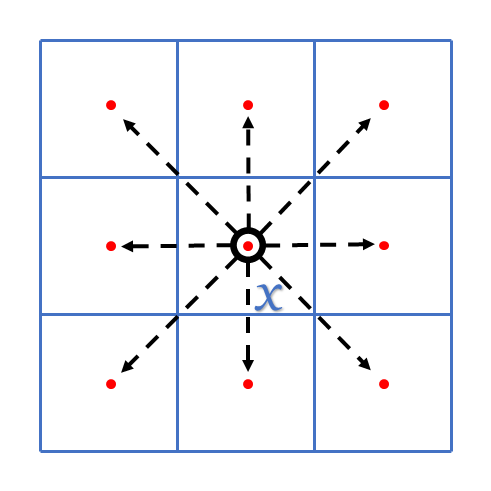} 
		\includegraphics[width = 0.45\linewidth]{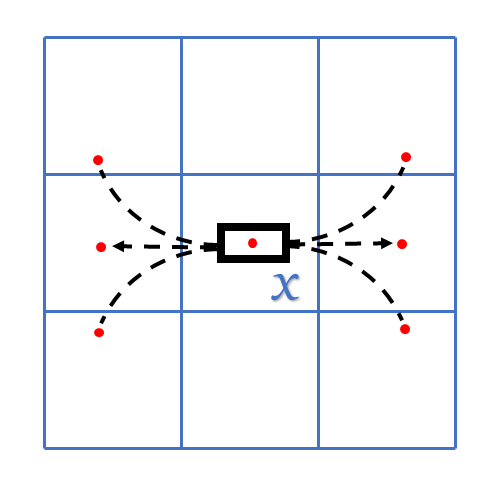} 
	} 		 
	\caption{\textbf{Left:} A* associates costs with centers of cells and only visits states that correspond to grid-cell centers. \textbf{Right:} Hybrid A* associates a continuous state with each cell, and the score of the cell is the cost of its associated continuous state. A* path can always move to the center of the adjacent node, while in hybrid A*, we consider the actual motion constraints of the object, so the red dot does not appear in the grid center. }
	\label{fig:a_star}
	\vspace{-3.5mm}
\end{figure}

A*~\cite{A} search can be seen as an improvement of Dijkstra's search. Dijkstra calculates the cost to start $g(x)$ of each vertex to determine the next vertex to be expanded. A* search enhances the algorithm by using heuristic cost $h(x)$, allowing faster convergence under certain conditions, while still ensuring its optimality \cite{hart1968formal}. The heuristic cost $h(x)$ is the cost to goal based on a heuristic estimate of the cost from state $x$ to the goal state $x_{goal}$, since the actual cost $g(x)$ is the path that has been actually traversed. The total cost is thus $$f(x) = g(x)+h(x)$$ by which the way points will be sorted. A standard heuristic estimate function is the Euclidean distance for two dimensional problems.

The hybrid A*\cite{dolgov2008practical, hybridA} algorithm is proposed for path planning of nonholonomic robots. In A*, we do not consider the direction of the moving object, and we do not consider the actual movement of the object. However, in hybrid A*, we need to consider the constraint of the robot motion model. In Figure \ref{fig:a_star}, we use the red dot to indicate the possible position of the robot. 
% In other words, the robot can only reach the position determined by the motion model in the next search \cite{dolgov2010path, montemerlo2008junior}. 
The differences between the two algorithms are shown in Table~\ref{table:difference}, and we also provide pseudo-code in Alg.~\ref{alg:hybrid_a_star}.

% For the implementation, both of the frameworks of hybrid A* and A* can be explained by the pseudo-code Alg.~\ref{alg:hybrid_a_star}. The differences are the cost update function $f(x) = g(x) + h(x)$ and neighbor search function $neig(x)$ in the pseudocode. In A*, the actual cost $g(x)$ is the length of the path. In hybrid A*, many other factors need to be considered, such as the change of the heading angle, whether to reverse the robot and so on. The heuristic cost $h(x)$ of hybrid A* takes the maximum of the following two items: 1) The A* cost from the current state to the end state. 2) The length Reeds-Shepp curves from the current state to the end state. 

\begin{table}[h]

% \Large
\begin{adjustbox}{max width=\columnwidth}

\begin{tabular}{|c|c|c|}
\hline
           & Hybrid A*                                            & A*                                   \\ \hline
Dimension & (x, y, $\theta$)                        & (x, y)                               \\ \hline
Vertex     & Possible movement paths & Grid map cells \\ \hline
g(x)       & Kinematic model                                      & Manhattan / Euclidean                \\ \hline
h(x)       & Max(Reeds\_Shepp Dist, A*)                                & Manhattan / Euclidean                \\ \hline
\end{tabular}

\end{adjustbox}
\caption{Differences between A* and Hybrid A*}
\label{table:difference}
\vspace{-10pt}
\end{table}

\begin{algorithm}[t]
	\KwIn{ Start state: $x_s$;  Goal state: $x_g$}
		
    \KwOut{ Valid path between $x_s$ and $x_g$ }
	
	\Begin{
	
	    $O$ = $\emptyset$  \noteCM{ // Initialize Open set }
	    
	    $C$ = $\emptyset$  \noteCM{ // Initialize Close set }
	    
	    $f(x_s) = g(x_s) + h(x)$  \noteCM{ // Update cost of $x_s$ according to the cost function. $g(x)$ is the actual cost and $h(x)$ is the heuristic cost.}
	    
	    $O.push(x_s)$
	    
	    \While{$O$ not empty}{
	        
	        $x$ $\leftarrow$ $O.popMin()$  \noteCM{ // $O.popMin()$ return the node with the lowest cost in O }
	        
	        \If{$x$ == $x_g$}{
	            return $path$  \noteCM{ // Trace the parent node from the end point $x$, until it reaches the starting point, return to the result $path$ found.}
	        } \Else {
	            $C.push(x)$
	            
	            \For{each $n \in neig(x)$}{  \noteCM{// Go through all collision-free neighbors of $x$ according to the kinematic model}
	            
    	            \If{$n \notin O$}{
    	                $f(n) = n.updateCost()$
    	                
    	                $O.push(n)$
    	            }
	            }
	        }
	    }
	    return $null$ \noteCM{// Can not find a valid path}
    } 
	\caption{Hybrid A* Search}
	\label{alg:hybrid_a_star}  
% 	\vspace{-6pt}
\end{algorithm}

\begin{figure}[t]
    \centering
    \vspace{-2pt}
    \includegraphics[width=0.9\columnwidth]{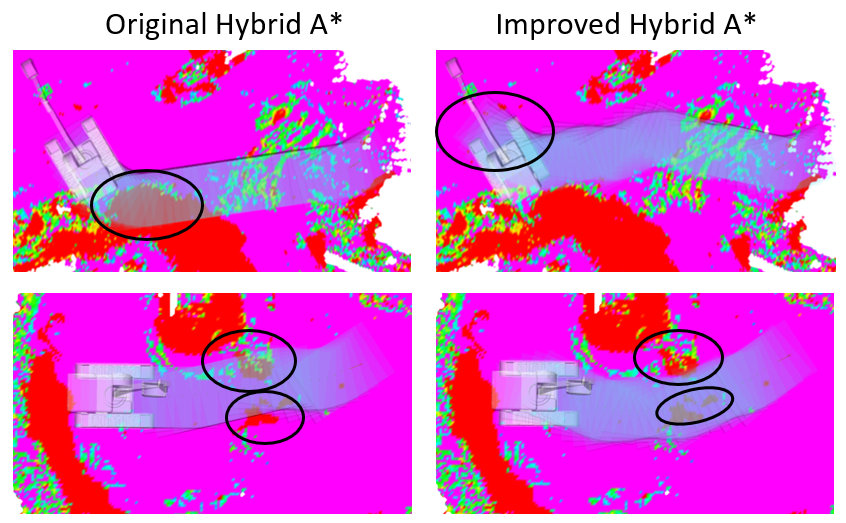}
    % \vspace{-4pt}
    \caption{\textbf{Comparison between original and improved hybrid A* planners.} Our planning method has more flexibility, including running over small obstacles between two tracks based on our traversability map.}
    \label{fig:planningfailed}
    % \vspace{-16pt}
    \vspace{-18pt}
\end{figure}

\subsection{Experiment Details of Offline Planning}

We compare the success rate of the planner using output traversability maps from the geometric-only method~\cite{geo1} and the proposed \ours{}. We use 9 different scenarios, and each scenario is tested with the same starting points and random goal position with more than 10 trials. We list the details of all scenarios in Table~\ref{tab:exp}.

\begin{table}[t]
\centering
\resizebox{0.9\columnwidth}{!}{%
\begin{tabular}{lcc|cc}
  \toprule
Scenarios & Difficult Terrain & Obstacles & Geometric Method~\cite{geo1} (\%)  & \ours{} (\%)  \\ 
  \midrule
Case 1   & \checkmark &  \checkmark    & 40      & \textbf{60  } \\
Case 2   & \checkmark &  \checkmark    & 16.67      & \textbf{71.43  } \\
Case 3   & \checkmark &      & 50      & 50   \\
Case 4   & \checkmark &  \checkmark    & 50      & \textbf{100  } \\
Case 5   & &  \checkmark    & 20      & \textbf{100  } \\
Case 6   & \checkmark &  \checkmark      & 40      & \textbf{80  } \\
Case 7   & &  \checkmark      & 50      & 50   \\
Case 8   & & \checkmark     & 20      & \textbf{60  } \\
Case 9   & \checkmark &      & 20      & \textbf{60  } \\
  \midrule
Overall & \checkmark & \checkmark & 33.3   & \textbf{82.6  } \\
\bottomrule
\end{tabular}
}
\caption{\textbf{Success rate of planning in each scenario.} ``Difficult Terrain'' means the excavator must traverse through or navigate around a rough region or water. ``Obstacles'' means there are obstacles in the environment.}
\label{tab:exp}
\vspace{-10pt}
\end{table}

\subsection{Failure Cases of Hybrid A* in Our Applications}

We show some planning scenarios where traditional Hybrid A* would fail in Fig~\ref{fig:planningfailed}.

\vspace{-10pt}
\section{More Visualization}

\subsection{Testing Site}

\begin{figure}[t]
    \centering
    \includegraphics[width=0.95\columnwidth]{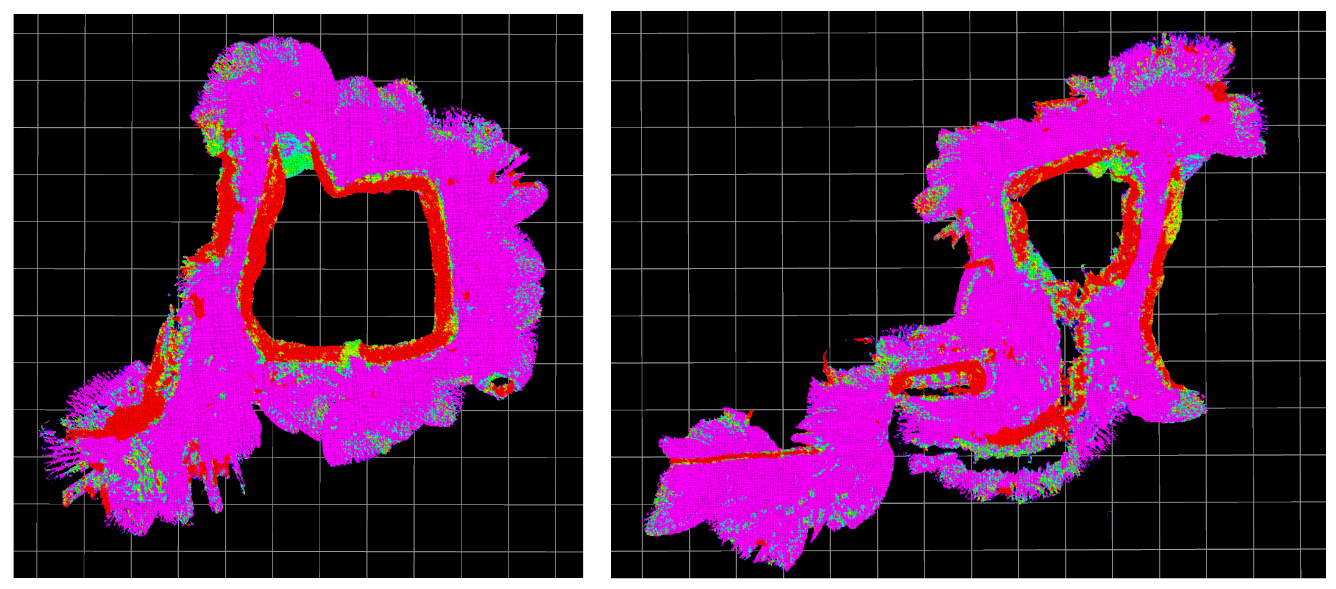}
    \caption{\textbf{Traversability maps of our two testing sites.} Each grid is 10 m by 10 m.}
    \label{fig:testing}
    % \vspace{-15pt}
\end{figure}

\begin{figure}[t]
    \centering
    % \vspace{-15pt}
    \includegraphics[width=0.95\columnwidth]{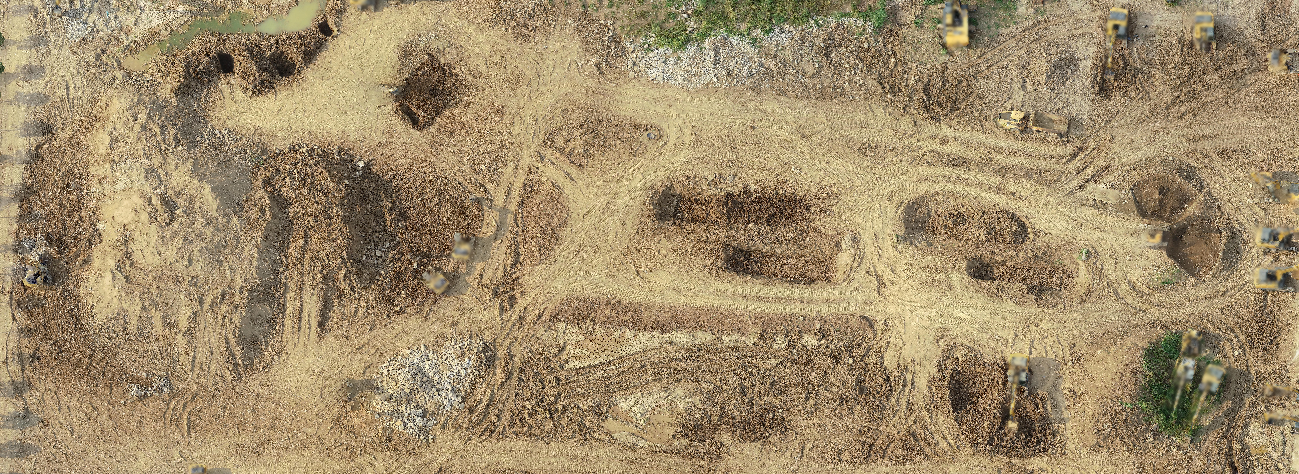}
    \caption{\textbf{Aerial image of our testing site taken in 2020.}}
    \label{fig:site}
    \vspace{-15pt}
\end{figure}

We show the traversability maps of two testing sites in Figure~\ref{fig:testing}. In Figure~\ref{fig:site}, we show a drone image taken in 2020. Note that the image is outdated, and the condition might be different from when our experiments are done.

\begin{figure}[t]
    \centering
    \includegraphics[width=0.8\columnwidth]{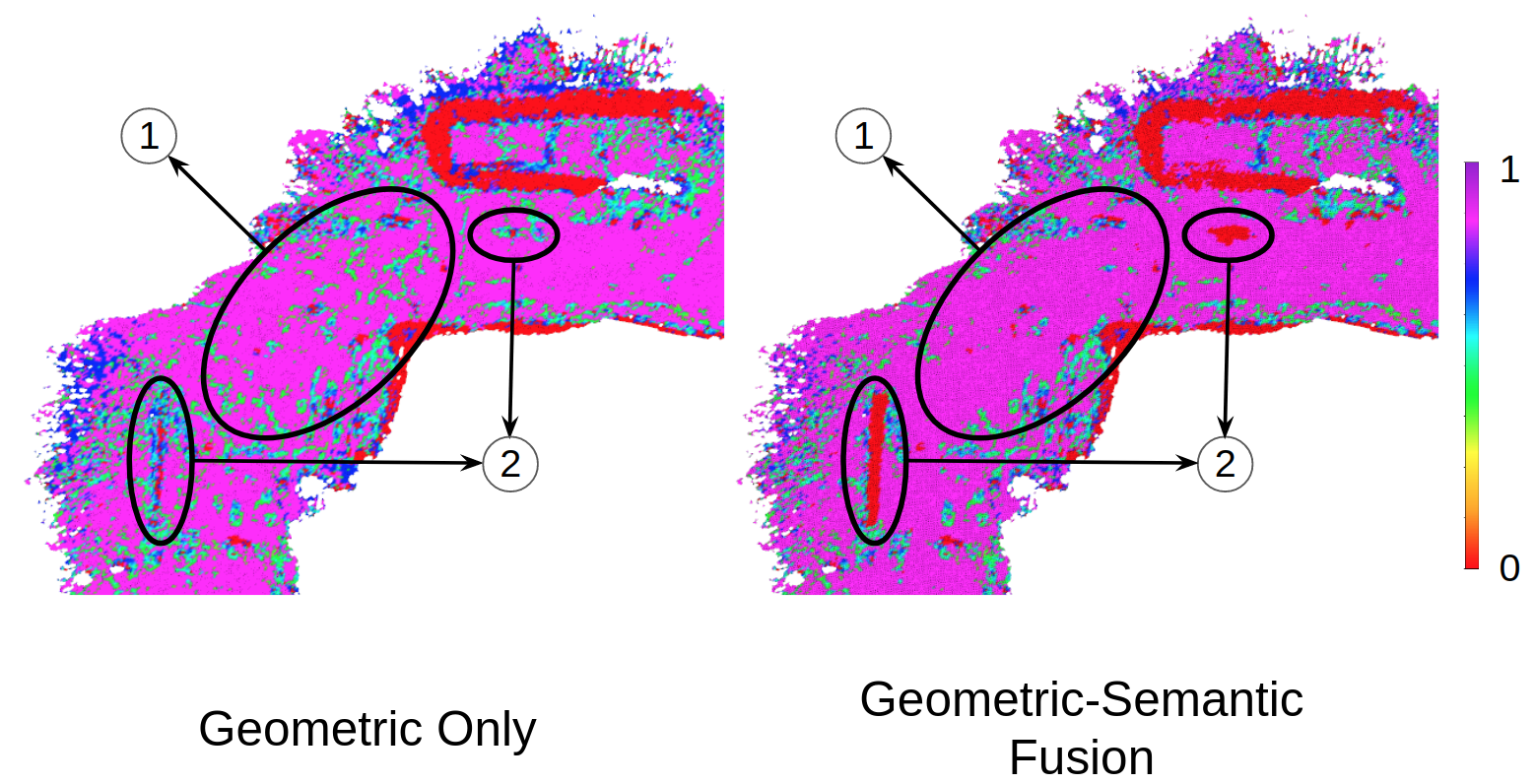}
    \caption{\textbf{Grid map comparison between the geometric-only scheme~\cite{geo1} and ours:} (1) Our method is less noisy and has more connected regions to plan a feasible trajectory. (2) Our method can detect obstacles that the geometric method could not recognize.}
    \label{fig:qualitative}
    % \vspace{-10pt}
\end{figure}

\subsection{Qualitative Comparisons on Mapping methods}

In Figure~\ref{fig:qualitative}, we compare traversability maps generated using a geometric-only method~\cite{geo1} and using \ours{} with geometric-semantic fusion. The output after fusion is less noisy since segmentation results can smooth out safe regions. Our method detects more non-traversable regions based on obstacles and dangerous regions from semantic information.

%%%%%%%%%%%%%%%%%%%%%

% \section*{Acknowledgments}

%% Use plainnat to work nicely with natbib. 

\bibliographystyle{plainnat}
\bibliography{citation}

\end{document}